\documentclass[]{spie}  

\usepackage{amsmath,amsfonts,amssymb}
\usepackage{graphicx}
\usepackage[colorlinks=true, allcolors=blue]{hyperref}
\usepackage{multirow}
\usepackage{multicol}
\usepackage[table]{xcolor}
\usepackage{tikz}

\title{Towards Improved Semiconductor Defect Inspection for high-NA EUVL based on SEMI-SuperYOLO-NAS}

\author[a,b,*]{Ying-Lin Chen}
\author[b,c,*]{Jacob Deforce}
\author[b,c,*]{Vic De Ridder}
\author[b, *]{Bappaditya Dey}
\author[b]{Victor Blanco}
\author[d, `]{Sandip Halder}
\author[b]{Philippe Leray}

\affil[a]{Faculty of Science, KU Leuven, 3001 Leuven, Belgium}
\affil[b]{Interuniversity Microelectronics Centre (imec), Kapeldreef 75, 3001 Leuven, Belgium}
\affil[c]{Faculty of Engineering, Ghent University, 9000 Ghent, Belgium}
\affil[d]{SCREEN SPE Germany GmbH, Ismaning, 85737, Germany}
\affil[*]{Equal Contribution}
\affil[`]{This research was conducted during Sandip Halder’s tenure at imec }

\authorinfo{Further author information: (Send correspondence to Bappaditya Dey)\\  Bappaditya Dey: E-mail: Bappaditya.Dey@imec.be}

\definecolor{custom_dark_blue}{RGB}{21,96,130}
\definecolor{custom_orange}{RGB}{233,113,50}
\definecolor{custom_dark_green}{RGB}{25,107,36}
\definecolor{custom_light_blue}{RGB}{15,158,213}

\begin{document} 

\maketitle
\begin{abstract}

Continuous progression of Moore’s law brings new challenges in metrology and defect inspection. As the semiconductor industry embraces High-Numerical Aperture Extreme Ultraviolet Lithography (High-NA EUVL), there is a current industry-wide evaluation of this technology for potential pitch reduction in future nodes. One of the primary hurdles in implementing High-NA EUVL in High Volume Manufacturing (HVM) is its low depth of focus. Consequently, suppliers of resist materials are compelled to opt for thin resist and/or new underlayers/hardmask’s. Experimental combinations of thin resist materials with novel underlayers and hardmask's seem to pose signal detection challenges due to poor Signal-to-Noise Ratio (SNR). In such a scenario, manual classification of these nano-scale defects faces limitations in terms of required time and workforce, and the robustness and generalizability of outcomes are also questionable. In recent years, vision-based machine learning (ML) algorithms have emerged as an effective solution for image-based semiconductor defect inspection applications. However, developing a robust ML model across various image resolutions without explicit training remains a challenge for nano-scale defect inspection. The goal of this research is to propose a scale-invariant Automated Defect Classification and Detection (ADCD) framework capable to upscale images, addressing this issue. We propose an improvised ADCD framework as \textbf{SEMI-SuperYOLO-NAS}, which builds upon the baseline YOLO-NAS architecture. This framework integrates a Super-Resolution (SR) assisted branch to aid in learning high-resolution (HR) features by the defect detection backbone, particularly for detecting nano-scale defect instances from low-resolution (LR) images. Additionally, the SR-assisted branch can recursively generate or reconstruct upscaled images ($\sim\times$\textbf{2}/$\times$\textbf{4}/$\times$\textbf{8}...) from their corresponding downscaled counterparts, enabling defect detection inference across various image resolutions without requiring explicit training. Moreover, we investigate improved data augmentation strategy aimed at generating diverse and realistic training datasets to enhance model performance. We have evaluated our proposed approach using two original FAB datasets obtained from two distinct processes and captured using two different imaging tools. Finally, we demonstrate zero-shot inference for our model on a new, originating from a process condition distinct from the training dataset and possessing different CD/Pitch characteristics. Our experimental validation demonstrates that our proposed ADCD framework aids in increasing the throughput of imaging tools ($\sim\times$\textbf{8}) for defect inspection by reducing the required image pixel resolutions.

\end{abstract}


\section{Introduction}
\label{sec:intro}  

\begin{figure}[!ht]
\centering
\includegraphics[width=8cm]{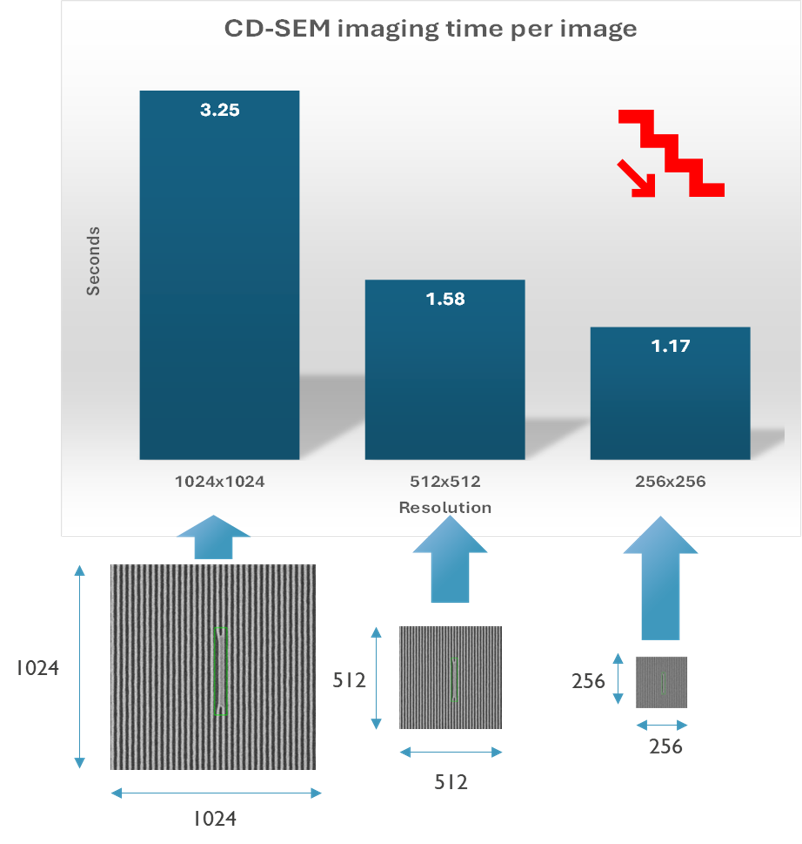}
\caption{Depiction of CD-SEM imaging time (sec.) against image resolution (WXH).}
\label{fig1_new}
\end{figure}

\begin{figure}[!ht]
\centering
\includegraphics[width=8cm]{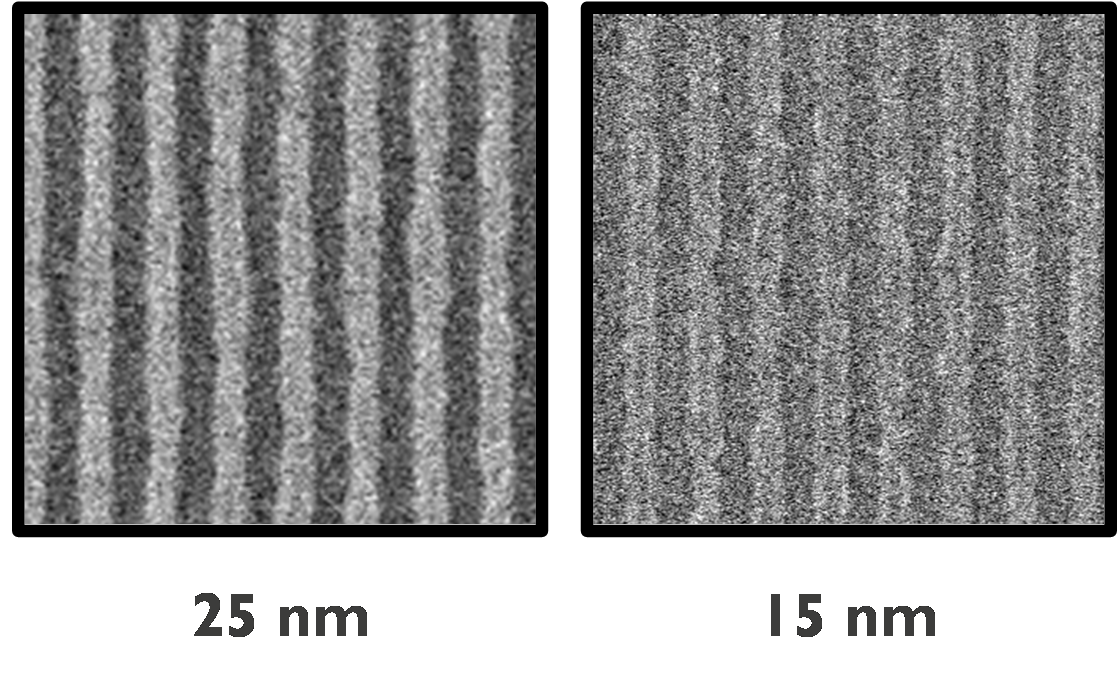}
\caption{CD-SEM images comparing resist thicknesses of 25nm and 15nm. As the resist thickness decreases, the SEM Signal-to-Noise Ratio (SNR) deteriorates.}
\label{intro1}
\end{figure}

At the advanced semiconductor nodes (towards sub-30nm pitches for 5nm node and below), the smallest possible pattern structures are targeted, accompanied by increased device complexity. To sustain this scaling trajectory, the industry is currently evaluating the utilization of High-Numerical Aperture Extreme Ultraviolet Lithography (High-NA EUVL)\cite{weiss2023overlay}. However, EUVL necessitates an exceptionally long inspection time due to its propensity to induce stochastic defects\cite{ouchi2020trainable}. To detect these nano-scale defects (if stochastic in nature) with a high level of certainty, Electron-beam (e-beam) tools typically need to inspect more areas on a wafer (generally an entire wafer)\cite{kondo2021massive}, commonly utilizing Scanning Electron Microscope (SEM) images. However, the e-beam inspection process itself suffers from low throughput and requires high-resolution SEM images. Conversely, SEM imaging throughput is contingent upon image resolution. This problem is depicted in Fig.\ref{fig1_new}, where we observe that the CD-SEM imaging time (in seconds) per image is directly proportional to the image resolution. For instance, acquiring a SEM image at a resolution of (1024x1204) takes nearly twice as long as acquiring the same image at a resolution of (512x512). Another key challenge in integrating High-NA EUVL into High Volume Manufacturing (HVM) is the low depth-of-focus (DoF). This factor is driving resist material suppliers to adopt thin(ner) resists\cite{lorusso2022metrology}, alongside innovative underlayers and hardmasks. However, the combination of thin resist materials with these novel underlayers and hardmasks often gives rise to signal detection challenges for inspection and metrology equipment, attributable to a low Signal-to-Noise Ratio (SNR), as illustrated in Fig.\ref{intro1}. 

Our proposed research concept is based on addressing several existing fundamental problems:
\begin{enumerate}
    \item[(a)] To improve e-beam inspection throughput: The potential to accelerate conventional e-beam inspection lies in the utilization of low-resolution (LR) images. However, this approach encounters challenges in effectively inspecting such images. 
    \item[(b)] To improve SEM imaging throughput:Two primary factors contribute to increased imaging time: 
    \begin{enumerate}
        \item[$<i>$]  the necessity for higher frame numbers for averaging to enhance Signal-to-Noise Ratio (SNR). However, this requires additional time and can be generally destructive for resist materials, leading to pattern CD (Critical Dimension) shrinkage. Nevertheless, our previous research introduced an unsupervised deep learning denoiser\cite{dey2021sem} aimed at enhancing the SNR of raw images for low frame numbers while simultaneously maintaining reliable roughness measurements for thin resist\cite{zidan2022extraction}, thereby improving CD-SEM imaging throughput.
        \item[$<ii>$] While a larger pixel size facilitates faster defect inspection within the same Field-Of-View (FOV), acquiring higher-resolution images demands more time due to the additional data and detail that need to be captured and processed. Conversely, lower pixel resolution leads to reduced detail, thus adversely impacting the detection of small defects. This trade-off is illustrated in Fig.\ref{fig1_new}.
    \end{enumerate}
\end{enumerate}

In the past, defect inspection was either conducted manually by human experts or relied on less effective rule/reference-based methods or algorithms. However, the challenges mentioned above in e-beam inspection of EUVL processed wafers have significantly degraded the performance of these rule-reference-based methods. With the emergence of computer vision-based advanced deep learning algorithms, particularly Convolutional Neural Networks (CNN)-based approaches, there is a growing popularity in semiconductor defect inspection applications, primarily because of their robustness and generalizability.
Our main objective in this research is to accelerate e-beam inspection primarily by utilizing low-resolution SEM images (addressing challenge (a)), and to enhance SEM imaging throughput(providing a solution for challenge (b) $<ii>$), thereby improving nano-scale defect detection performance.

This research work contributes in the following ways:
\begin{enumerate}
    \item We propose an improvised ADCD framework as SEMI-SuperYOLO-NAS, integrating a SR-assisted branch to facilitate high-resolution (HR) feature learning for detecting nano-scale defects in LR-images.
    \item We demonstrate a scale-invariant ADCD framework capable of upscaling images from their low-resolution counterparts, enabling defect inspection across different image resolutions without explicit training. This facilitates increased image acquisition throughput (\#Frames per second/\#images per second) for imaging tools.
    \item We investigate an improved data augmentation strategy to create diverse and realistic training datasets, leading to a more accurate and robust ADCD framework.
    \item We benchmark the proposed framework on two original FAB datasets, as SEM-ADI and EDR-AEI. Furthermore, we demonstrate the "Zero-Shot" inference capability by making predictions on defect classes not seen during training, utilizing "(weakly) semantic embeddings."
\end{enumerate}

 In Section 2, we will review related research works on semiconductor defect inspection. Section 3 will delve into the details of the proposed methods and experiments conducted. Sections 4 and 5 will present our results and conclusions.

\section{Related Work}

In this section, we review various works related to defect inspection in the semiconductor industry employing machine learning techniques.  Chang et al.\cite{chang2005unsupervised} proposed an unsupervised self-organizing neural network to address challenges in post-sawing wafer defect inspection, previously reliant on human inspection, thus requiring significant labor force and suffering from human fatigue. Dey et al.\cite{dey2022deep} utilized a ResNet architecture-based ensemble model for nano-scale defect inspection and classification. An unsupervised machine learning model was applied to reduce false-positive defects arising from SEM noise. Dey et al. also benchmarked another ensemble defect detection framework based on the state-of-the-art one-stage object detection model YOLOv5\cite{dey2022towards}, which outperformed the previous ResNet-based ensemble framework.

A data-centric approach was also demonstrated\cite{dehaerne2023yolov8}, showing a way to consensually ensemble labels from different anonymous labelers for the same dataset, requiring minimal intervention from experts for defect labeling expectations. The final combined label, along with expert-aware post-processing data, yielded the highest test mAP of 0.919 with the state-of-the-art YOLOv8 model. The concept of pseudo-label prediction and model ensembles from different labelling partitions to avoid duplicate labeling work was proposed. Cheon et al.\cite{cheon2019convolutional} utilized a Convolutional Neural Network (CNN) along with a k-NN method to classify wafer surface defects from SEM images. The final activation layer outputs were used as features for k-Nearest Neighbors (k-NN), but with a threshold for clustering, allowing the algorithm to identify unknown defect classes. However, when a new class is discovered, the CNN needs retraining to ensure consistent high performance over time. To address this issue, the authors proposed the idea of incremental training. In previous study\cite{rahman2023noise}, a noise characterization for future denoising algorithms demonstrated that as scanning speed increased, a shift in noise property from a Gaussian to a Gamma distribution was observed. In recent years, CD-SEM tool vendors have also incorporated machine learning to enhance tool efficiency\cite{ouchi2020trainable,kondo2021massive}. The idea of ML-based super-resolution defect inspection was proposed\cite{kondo2021massive}, where a CNN model takes the database of designed features and CD-SEM images as input and generates the pixel distribution for defect prediction. Their model utilized non-defect regions to train the algorithm, generating the pixel luminosity distribution for the given database. If the pixel brightness values of the original image significantly deviated from the predicted pixel distribution center, the pixel was classified as defective. This approach demonstrated the potential for identifying bridge defects and necking defects even in lower pixel resolution images of 4nm/pixel. However, in this method, knowledge of the design database (mask layout) is necessary, making automated defect type classification challenging.  Nonetheless, their research highlights the advantages of massive metrology, showcasing how a FOV of 80$\times$80 microns can facilitate automatic defect identification through machine learning. Thus, the development of a super-resolution ML method capable of both identifying and classifying defects without requiring knowledge of the mask design remains a crucial missing component.

\section{Methodology} 

In this section, we provide a brief introduction to our training datasets, the proposed data augmentation strategy, an overview of the architecture of our proposed SEMI-SuperYOLO-NAS, and the strategy for preparing the dataset of High-resolution/Low-resolution (HR/LR) image pairs to train our SR-branch assisted defect detector model.

\subsection{Dataset}

\begin{figure}[ht]
\centering
\begin{minipage}{0.24\textwidth}
\centering
\includegraphics[width=\textwidth]{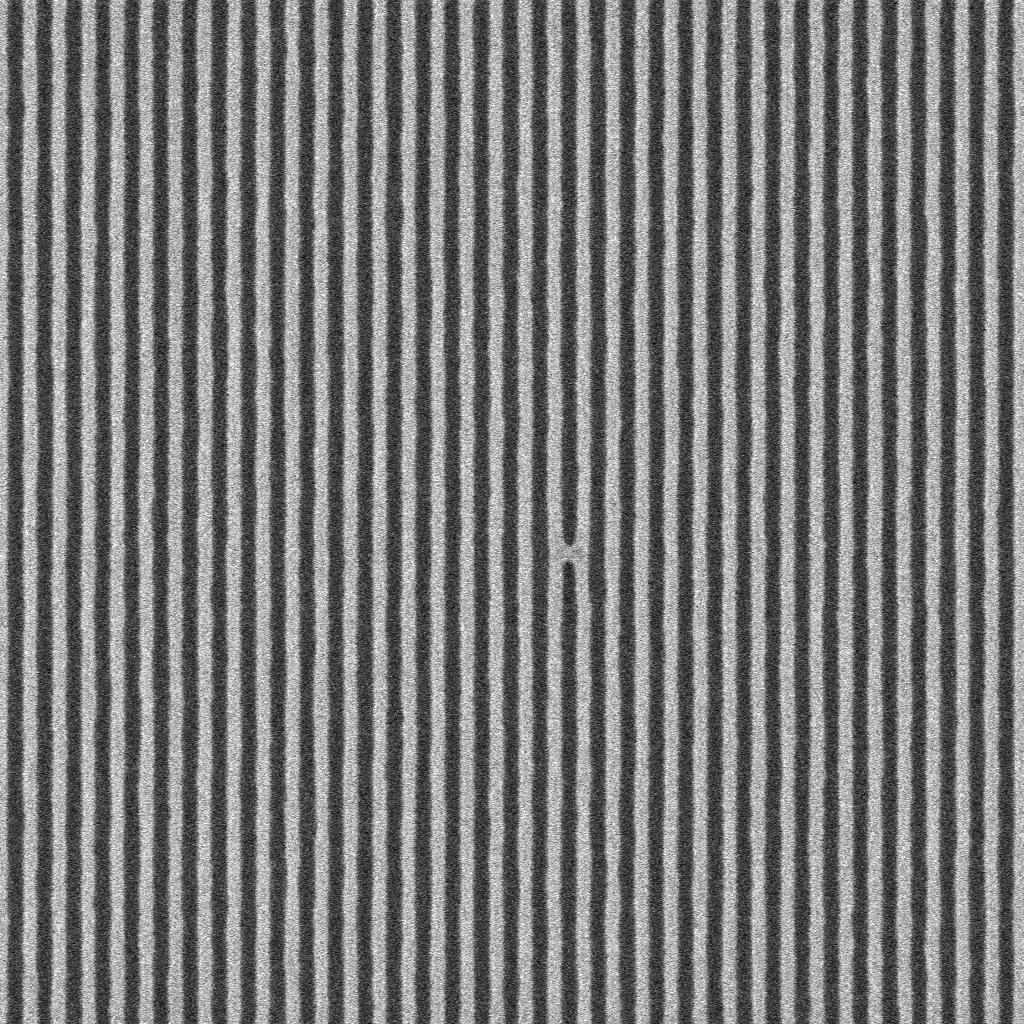}
(a)
\end{minipage}%
\hfill
\begin{minipage}{0.24\textwidth}
\centering
\includegraphics[width=\textwidth]{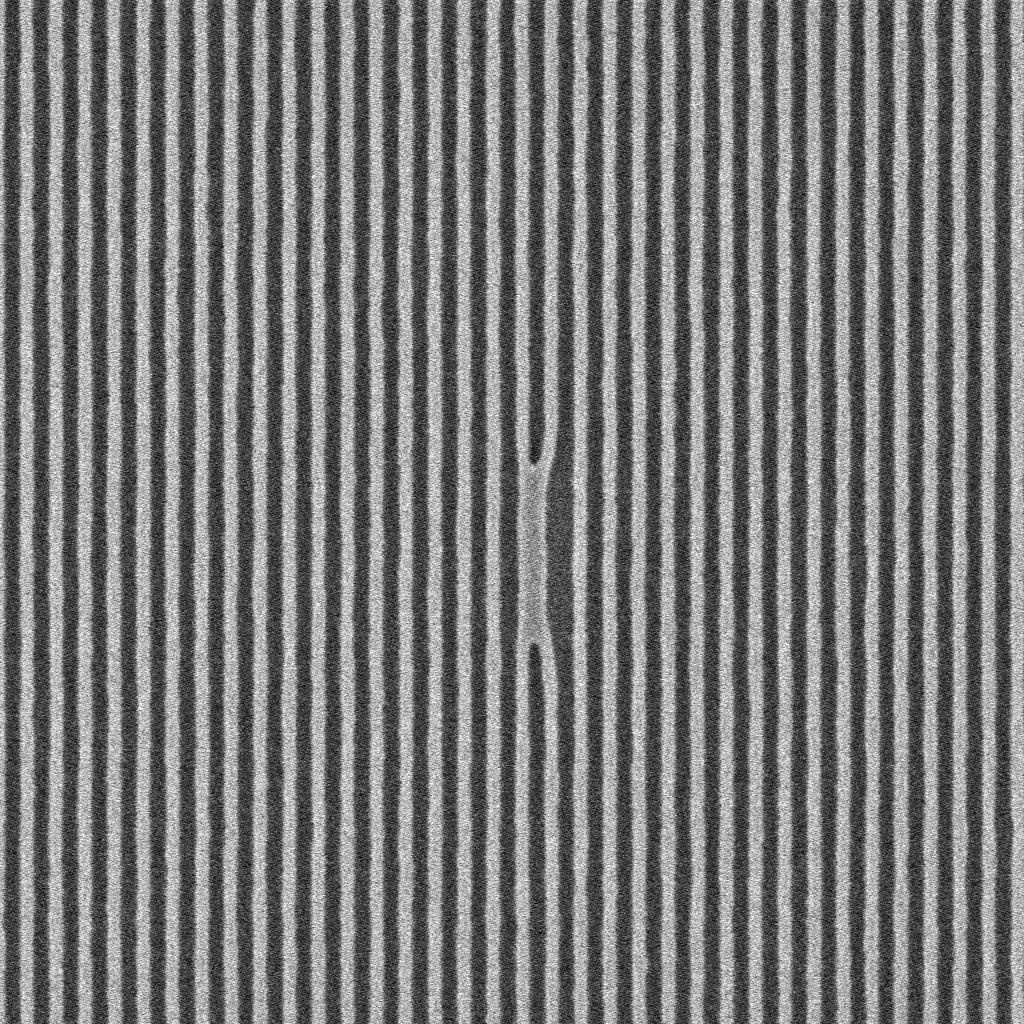}
(b)
\end{minipage}%
\hfill
\begin{minipage}{0.24\textwidth}
\centering
\includegraphics[width=\textwidth]{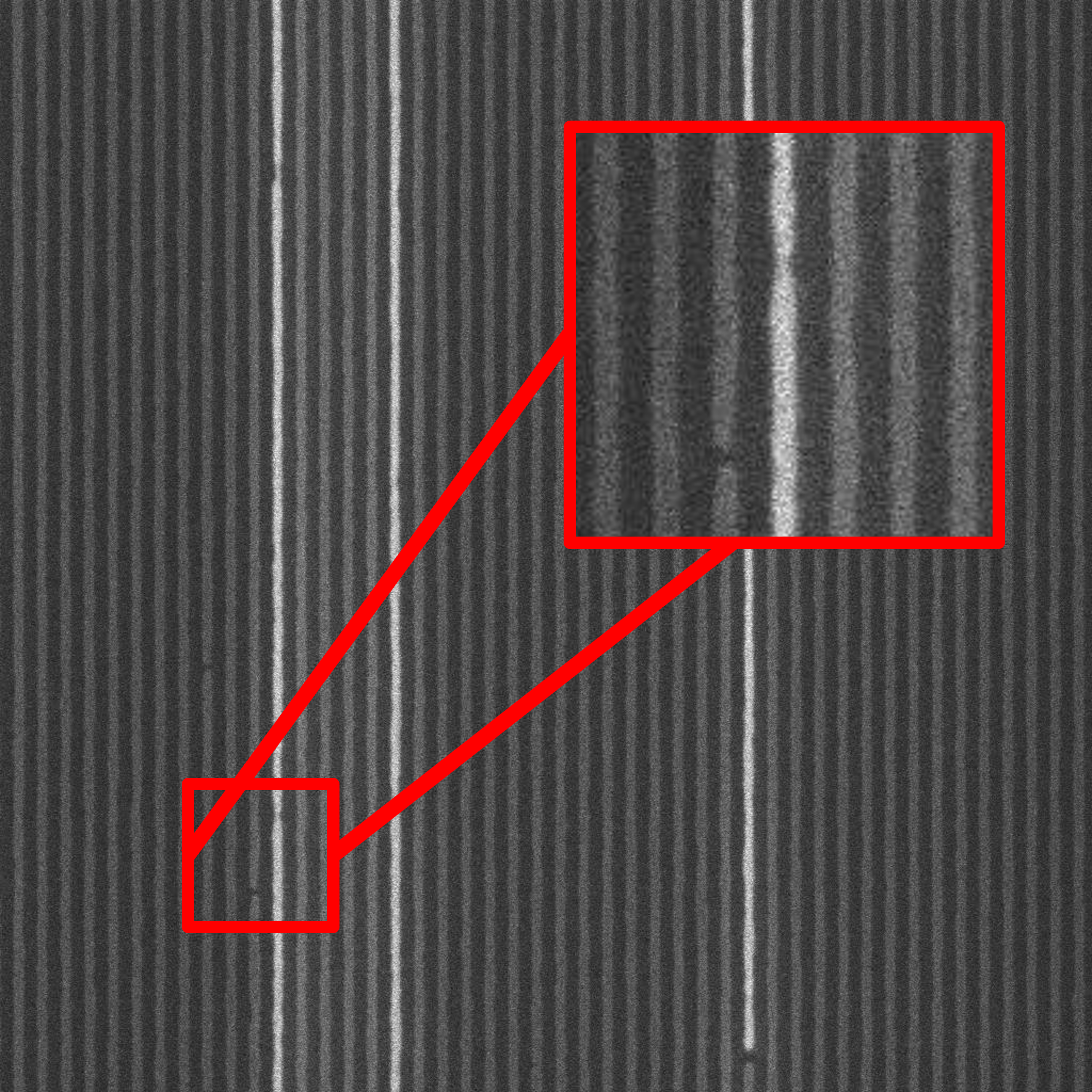}
(c)
\end{minipage}%
\hfill
\begin{minipage}{0.24\textwidth}
\centering
\includegraphics[width=\textwidth]{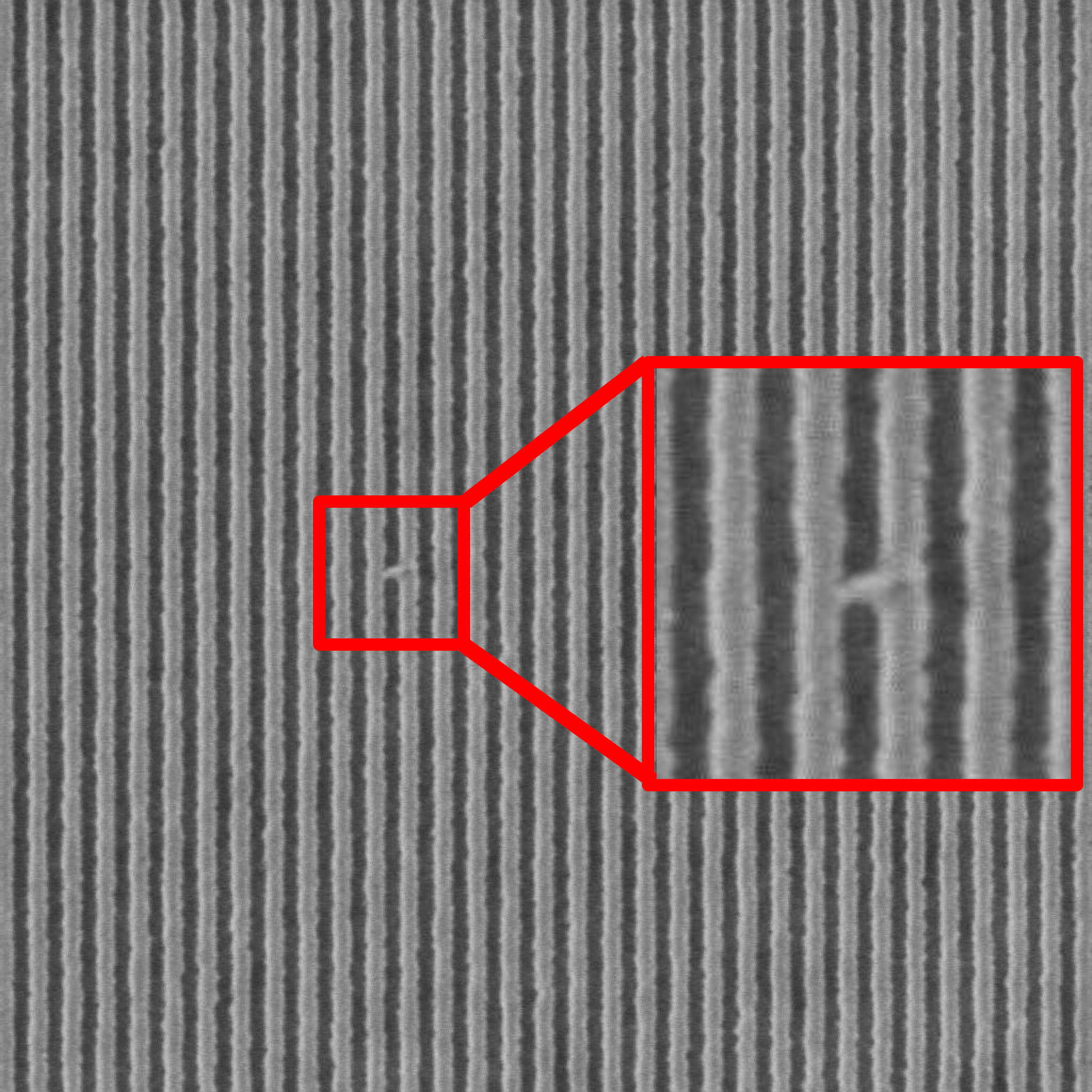}
(d)
\end{minipage}%
\vspace{4pt}
\caption{Defect class examples in SEM-ADI dataset. From left to right: (a) Bridge, (b) Line-collapse, (c) Gap \& Probable-gap and (d) Micro-bridges.}
\label{defect}
\end{figure}

\begin{figure}[ht]
\centering
\begin{minipage}{0.24\textwidth}
\centering
\includegraphics[width=\textwidth]{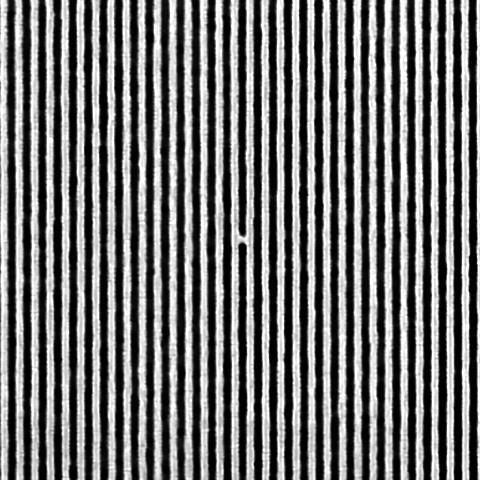}
(a)
\end{minipage}%
\hfill
\begin{minipage}{0.24\textwidth}
\centering
\includegraphics[width=\textwidth]{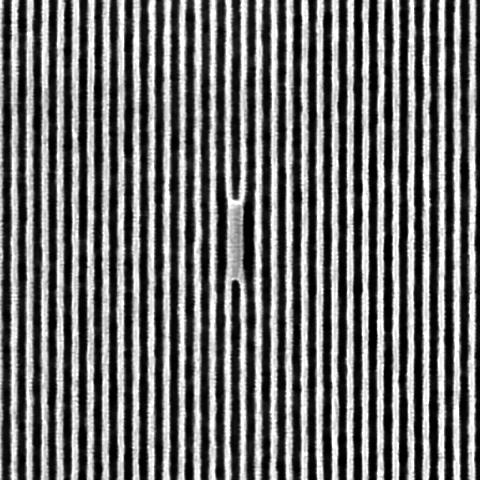}
(b)
\end{minipage}%
\hfill
\begin{minipage}{0.24\textwidth}
\centering
\includegraphics[width=\textwidth]{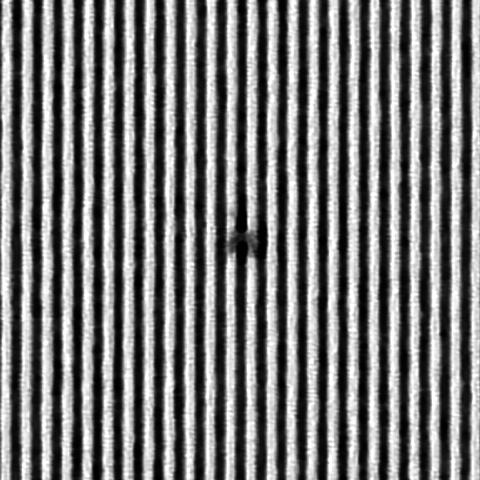}
(c)
\end{minipage}%
\hfill
\begin{minipage}{0.24\textwidth}
\centering
\includegraphics[width=\textwidth]{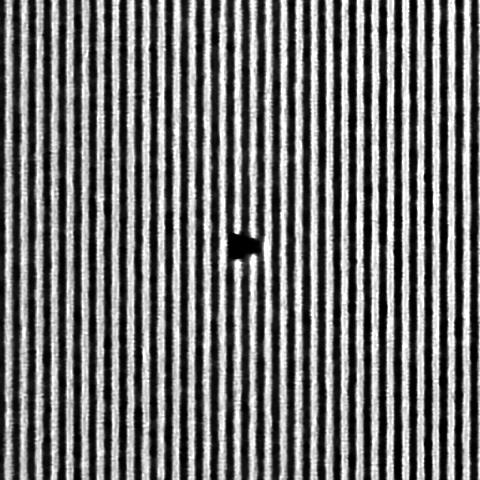}
(d)
\end{minipage}%
\vspace{4pt}
\caption{Defect class examples in EDR-AEI dataset. (a) bridge (b) pattern-collapse (c) dark-spot and (d) break.}
\label{AEI_defect}
\end{figure}

Two Line-Space (LS) pattern datasets were utilized. The first dataset, SEM-ADI, comprises 1324 images captured with the CD-SEM tool during After-Development-Inspection (ADI). The second dataset, EDR-AEI, comprises 527 images captured with the EDR tool during After-Etching-Inspection (AEI). Each image in these datasets contains at least one defect. SEM-ADI and EDR-AEI have original pixel resolutions of 1024$\times$1024 and 480$\times$480, respectively. SEM-ADI encompasses 5 different defect types, while EDR-AEI has 4 different defect types. We divided the two datasets into training and validation sets, as outlined in Table \ref{table1} and Table \ref{table2}, where we also provide details regarding the original defect classes and their respective total number of instances. In Fig.\ref{defect}, we illustrate the various defect types in the SEM-ADI dataset. In Fig.\ref{AEI_defect}, we present all defect types in the EDR-AEI dataset. The imbalance in defect classes and the limited availability of data in both datasets prompted us to develop novel augmentation strategies tailored to SEM and EDR imaging characteristics and assess their impact on model performance.

\begin{table}[htbp]
\vspace{10pt}
    \caption{SEM-ADI dataset with corresponding defect class distribution (Train/Val)}
    \vspace{2pt}
    \centering
    \begin{tabular}{|c|c|c|}
        \hline
         \textbf{Class Name} & \textbf{Train (1054 images)} & \textbf{Val (272 images)} \\
        \hline
         micro-bridge & 380 & 125 \\
        \hline
         gap & 1046 & 330 \\
        \hline
         bridge & 238 & 36 \\
        \hline
         line-collapse & 550 & 142 \\
        \hline
         probable gap & 315 & 103 \\
        \hline
    \end{tabular}

    \label{table1}
\end{table}

\begin{table}[!htbp]
\vspace{10pt}
    \caption{EDR-AEI dataset with corresponding defect class distribution (Train/Val)}
    \vspace{2pt}
    \centering
    \begin{tabular}{|c|c|c|}
        \hline
        \textbf{Class Name} & \textbf{Train (425 images)} & \textbf{Val (104 images)} \\
        \hline
         break & 126 & 28 \\
        \hline
         bridge & 107 & 27 \\
        \hline
         dark spot & 255 & 60 \\
        \hline
       pattern collapse & 137 & 34 \\
        \hline

    \end{tabular}

    \label{table2}
\end{table}

\subsection{Proposed  Data-augmentation strategy}
We implemented different image augmentation\cite{tvetkova2023robustness}  techniques to enhance the robustness of our model and address imbalances in the number of instances across defect classes. Within our image augmentation pipeline, utilizing the Albumentations library\cite{buslaev2020albumentations}, we integrated the following enhancements to introduce variability in illumination, noise, and atmospheric conditions:

\begin{itemize}
    \item Random Shadow: Generates sharp shadows of a random polygon shapes across the image.
    \item GaussNoise: Applies per-pixel-gaussian-distributed grayscale noise.
    \item RandomFog: Introduces blurry spots randomly throughout the image.
    \item RandomBrightness: Randomly changes the brightness and contrast of the input image.
    \item RandomGamma: Modifies image luminance using gamma correction to simulate different lighting conditions. 
    \item RingingOvershoot: Convolves an image with a 2D sinc filter to create a typical ringing artefact obtained when processing a bandlimited signal. Subtle edges appear around sharp color changes.

\end{itemize}

\begin{figure}[!ht]
\centering

\begin{minipage}{0.12\linewidth}
\centering
\includegraphics[width=\linewidth]{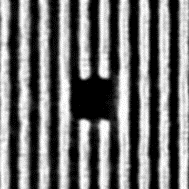}
\end{minipage}%
\hfill
\begin{minipage}{0.12\linewidth}
\centering
\includegraphics[width=\linewidth]{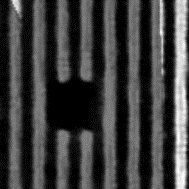}
\end{minipage}%
\hfill
\begin{minipage}{0.12\linewidth}
\centering
\includegraphics[width=\linewidth]{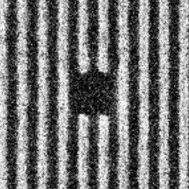}
\end{minipage}%
\hfill
\begin{minipage}{0.12\linewidth}
\centering
\includegraphics[width=\linewidth]{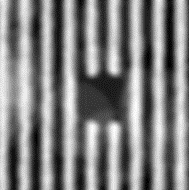}
\end{minipage}%
\hfill
\begin{minipage}{0.12\linewidth}
\centering
\includegraphics[width=\linewidth]{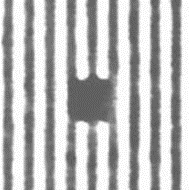}
\end{minipage}%
\hfill
\begin{minipage}{0.12\linewidth}
\centering
\includegraphics[width=\linewidth]{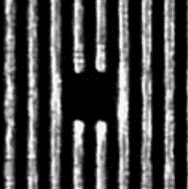}
\end{minipage}%
\hfill
\begin{minipage}{0.12\linewidth}
\centering
\includegraphics[width=\linewidth]{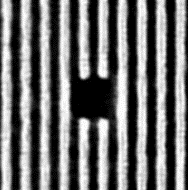}
\end{minipage}%
\hfill
\begin{minipage}{0.12\linewidth}
\centering
\includegraphics[width=\linewidth]{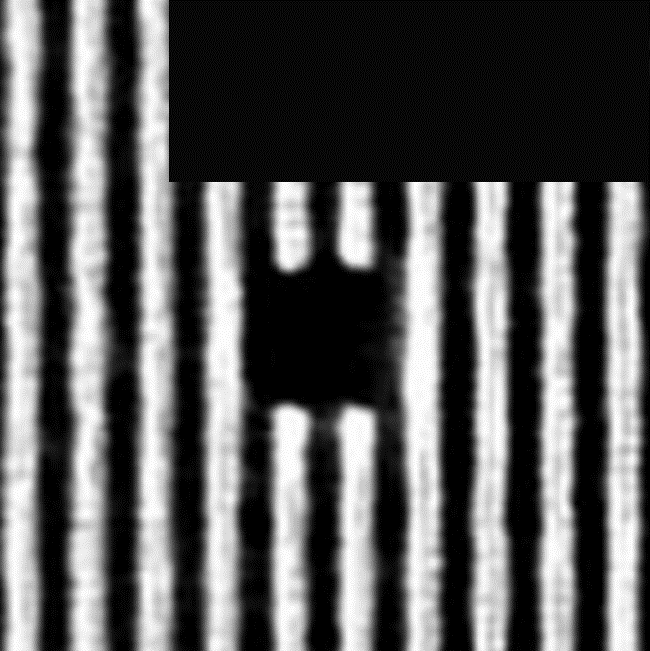}
\end{minipage}%
\vspace{4pt}
\hfill
\begin{minipage}{0.12\linewidth}
\centering
(a)
\end{minipage}%
\hfill
\begin{minipage}{0.12\linewidth}
\centering
(b)
\end{minipage}%
\hfill
\begin{minipage}{0.12\linewidth}
\centering
(c)
\end{minipage}%
\hfill
\begin{minipage}{0.12\linewidth}
\centering
(d)
\end{minipage}%
\hfill
\begin{minipage}{0.12\linewidth}
\centering
(e)
\end{minipage}%
\hfill
\begin{minipage}{0.12\linewidth}
\centering
(f)
\end{minipage}%
\hfill
\begin{minipage}{0.12\linewidth}
\centering
(g)
\end{minipage}%
\hfill
\begin{minipage}{0.12\linewidth}
\centering
(h)
\end{minipage}%
\hfill

\vspace{6pt}

\caption{Sample of augmentation on EDR-AEI break defect. (a) original image, (b) RandomShadow, (c) GaussNoise, (d) RandomFog, (e) RandomBrightness, (f) RandomGamma, (g) RandomOvershoot, and (h) Cutout.}
\label{aug_img}
\end{figure}

Additionally, we integrated a Cutout augmentation\cite{devries2017improved}, which involves removing a portion of the image. Moreover, we introduced two novel augmentations specifically for Line-Space pattern SEM images. Fig.\ref{aug_img} illustrates  the above-mentioned augmentation methods on the EDR-AEI dataset.

\begin{itemize}
    \item Copy defect: Replicates defect instances in different locations within the image. A rule-based algorithm aligns the pattern of the copied slice with the background, and its edges are faded to avoid any artefacts.
    \item ContrastChange: We sporadically decreased the horizontal darkness and contrast of the image. This mimics the effect observed in the SEM-ADI dataset.
\end{itemize}

\begin{figure}[ht]
\centering
\begin{minipage}{0.24\textwidth}
\centering
\includegraphics[width=\textwidth]{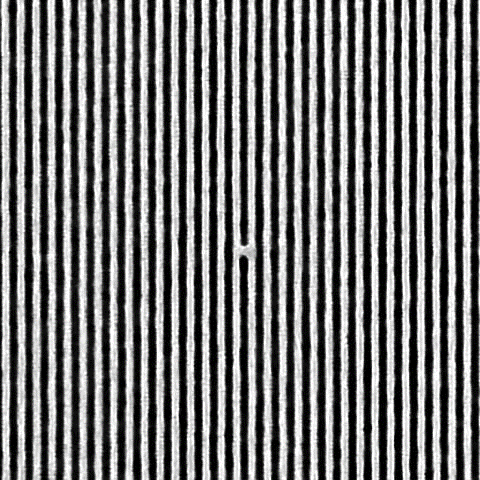}
(a)
\end{minipage}%
\hfill
\begin{minipage}{0.24\textwidth}
\centering
\includegraphics[width=\textwidth]{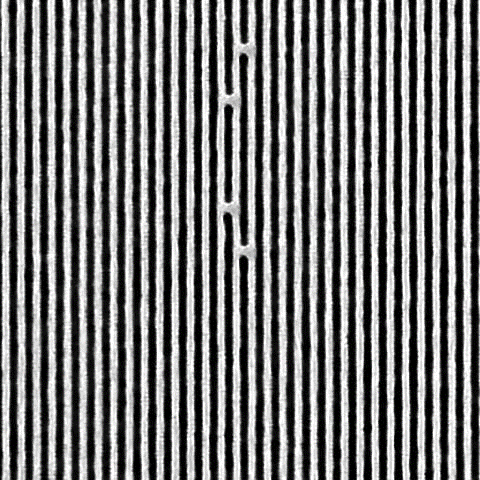}
(b)
\end{minipage}%
\hfill
\begin{minipage}{0.24\textwidth}
\centering
\includegraphics[width=\textwidth]{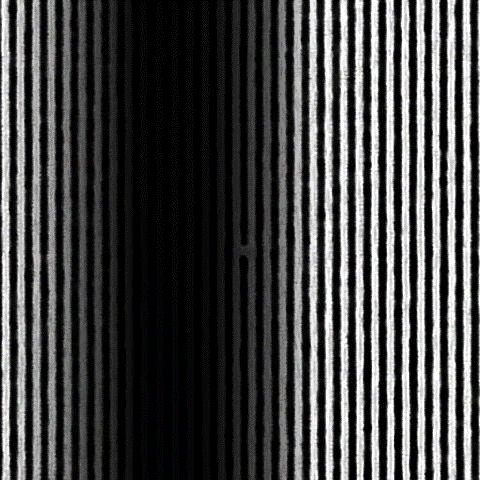}
(c)
\end{minipage}%
\hfill
\begin{minipage}{0.24\textwidth}
\centering
\includegraphics[width=\textwidth]{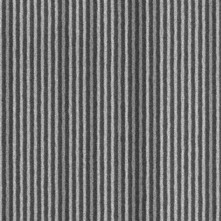}
(d)
\end{minipage}%
\vspace{4pt}
\caption{(a) original bridge defect (b) Copy Defect augmentation (c) ContrastChange augmentation and (d) orginal contrast change occurrence in SEM-ADI image.}
\label{aug_copy}
\end{figure}

We exclusively apply augmentation solely to the training dataset, aiming to further enhance the model's comprehension of semantic embeddings associated with defect types. During validation, we only use the original validation dataset, abstaining from any augmentation. The two novel augmentations tailored for LS defects are shown in Fig.\ref{aug_copy}.

As discussed in the preceding section,  we strategically apply augmentations on different defect types. For defect types that are easily discernible or already possess high instance numbers, we employ fewer augmentations. Conversely,  for defect types that are sparse in the original data or pose greater challenges for ADCD, we deliberately apply more augmentations. This approach reallocates training efforts towards addressing challenging tasks, while not over-allocating resources to defect types that are already reliably detected. The original number of defect instances, augmented defect instances, and the total number of instances after augmentation are detailed in Table \ref{aug_table}.

\begin{table}[htbp]
\vspace{8pt}
    \caption{Summary of Defect Instances in Original and Augmented Training Datasets.}
    \centering
    \begin{tabular}{|c|c|c|c|c|}
        \hline
        \multirow{2}{*}{\textbf{Dataset}} & \multirow{2}{*}{\textbf{Defect type}} & \textbf{\# of instances in} & \textbf{\# of instances in} & \textbf{total \# of } \\
        & & \textbf{original images} & \textbf{augmented images} & \textbf{instances} \\
        \hline
        \multirow{5}{2cm}{\centering \footnotesize \textbf{SEM-ADI}} &  micro-bridge &  380 & 350 & 730 \\
        \cline{2-5}
        & gap & 1046 & 44 & 1090\\
        \cline{2-5}
        & bridge &238 &117 & 355\\
        \cline{2-5}
        & line-collapse & 550 & 44 & 594\\
        \cline{2-5}
        &  probable gap & 315 & 714 & 1029\\
        \hline
        
        \multirow{4}{2cm}{\centering \footnotesize \textbf{EDR-AEI}} & break & 126 & 451 & 577 \\
        \cline{2-5}
        & bridge &107 & 450 & 557\\
        \cline{2-5}
        & dark spot & 255 & 902 & 1157\\
        \cline{2-5}
        &  pattern-coll. &137 & 360 & 497\\
        \hline
    \end{tabular}

    \label{aug_table}
\end{table}

\subsection{Dataset preparation for training SR-assisted branch}

\begin{figure}[!ht]
\vspace{10pt}
\centering
\includegraphics[width=10cm]{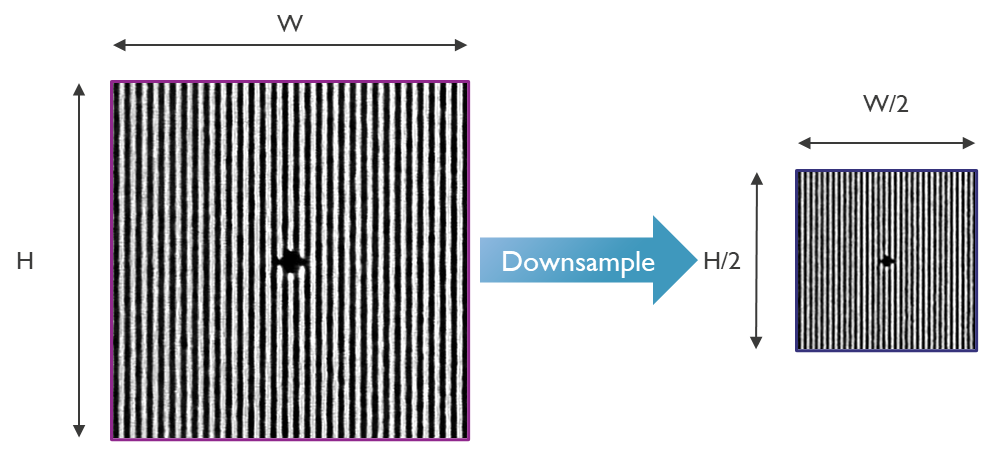}
\caption{Downsampling example: Bilinear Interpolation method applied to EDR-AEI Image to generate LR (240$\times$240) from HR (480$\times$480).}
\label{downsample}
\end{figure}

To facilitate training of SuperYOLOv5\cite{zhang2023superyolo} and proposed SEMI-SuperYOLO-NAS, we prepared datasets comprising pairs of High-resolution (HR) and Low-resolution (LR) images. Images were initially downsampled using the Bilinear interpolation method before any augmentation, as discussed in Section 3.2. For the SEM-ADI dataset, we downsampled the original resolution of (1024$\times$1024) to consecutive low-resolutions of (512$\times$512), (256$\times$256), and finally (128$\times$128). Similarly, for the EDR-AEI dataset, we downsampled the original image resolution of (480$\times$480) to (240$\times$240) and finally (120$\times$120). Fig.\ref{downsample} illustrates an example of downsampling for the EDR-AEI dataset. To train the SR-branch utilized by Super-YOLOv5\cite{zhang2023superyolo} and SEMI-SuperYOLO-NAS, higher resolution (HR) images were paired with corresponding lower resolution (LR) images as [{HR/LR} $\longrightarrow${1024/512}, {512/256}, {256/128}]. Additionally, baseline YOLO-NAS models were trained individually.

\subsection{Proposed SEMI-Super-YOLO-NAS architecture}

\begin{figure}[!ht]
\centering
\includegraphics[width=12cm]{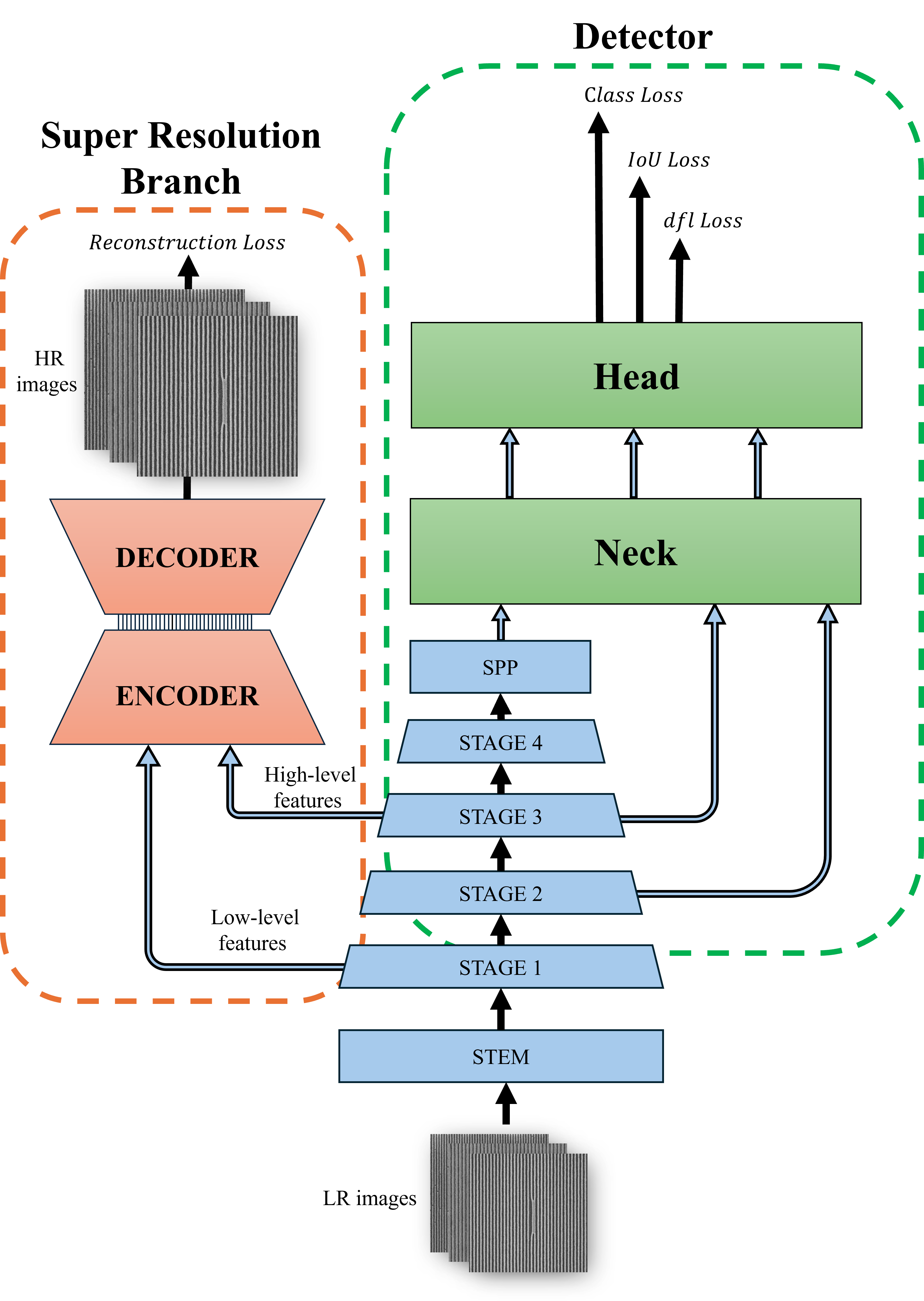}
\caption{Proposed SEMI-Super-YOLO-NAS architecture. }
\label{architecture}
\end{figure}

Our proposed SEMI-SuperYOLO-NAS architecture builds upon the recent baseline YOLO-NAS object detection architecture\cite{yolo-nas}. Neural Architecture Search (NAS)\cite{elsken2019neural} enables automatic exploration of a search space comprising candidate neural network architectures tailored for specific applications, such as object detection, guided by a search algorithm. This process involves sampling numerous architectures from the search space and evaluating them to select the most optimal configuration, thus replacing conventional neural network design methods typically performed by human experts. Another advantage of the YOLO-NAS architecture is its quantization-aware nature, which allows for striking a balance between information loss and latency. This contrasts with conventional object detection models, which often sacrifice precision in model weights to achieve a similar balance. The baseline YOLO-NAS utilizes the QARepVGG\cite{chu2024make} building block alongside the NAS algorithm to enhance accuracy preservation post-quantization. Fig.\ref{architecture} depicts the proposed SEMI-SuperYOLO-NAS architecture based ADCD framework. Zhang et al.\cite{zhang2023superyolo} proposed adding a Super-Resolution (SR) branch to the customized YOLOv5 architecture\cite{terven2023comprehensive}. They replaced the Focus module with a Multimodal Fusion module to prevent resolution degradation and spatial information loss, particularly for small object detection tasks. The SR structure consisted of a simple Encoder-Decoder model. This branch takes two backbone layers as input (one containing high-level features and the other containing low-level features) and outputs a high-resolution (HR) version of the original input image. Consequently, the loss function includes an additional term: the absolute difference between the reconstructed super-resolution image and the original HR image. This super-resolution loss function term facilitates learning HR feature extraction by the backbone, leading to improved detection performance, as demonstrated by Zhang et al.\cite{zhang2023superyolo} in their original work.

Inspired by this approach, we incorporated a Super-Resolution branch into the baseline YOLO-NAS’s architecture variants. We adopted Zhang et al.'s\cite{zhang2023superyolo} design for the SR-CNN architecture, which was specifically designed to minimize the impact on training time while still facilitating learning by the backbone. As mentioned earlier, this SR branch should take input from two locations within the YOLO-NAS-s backbone:

\begin{itemize}
    \item One where \underline{low-level features} have been extracted. For this purpose, we selected the output of the first backbone stage.
    \item One where \underline{high-level features} have been extracted. Here, we utilized the output of the third backbone stage.
\end{itemize}

Finally, we recursively train SEMI-SuperYolo-NAS on both SEM-ADI and EDR-AEI datasets, as :

\begin{enumerate}
    \item SR-assisted branch with $(x_{LR}^i, y_{HR}^i)$ image pairs to facilitate high-resolution features learning by the defect detection backbone,
    \begin{itemize}
        \item $x_{LR}\longrightarrow$ input downscaled image (say, 512, 256, 128,...)
        \item $y_{LR}\longrightarrow$ corresponding output super resolution/upscaled images (say, 1024, 512, 256,...)
    \end{itemize}
    \item Defect detection branch with $(x_{LR}^i, a_{\text{bbox}}^i)$,
    \begin{itemize}
        \item $x_{LR}\longrightarrow$ input downscaled image (say, 512, 256, 128,...)
        \item $a_{\text{bbox}}\longrightarrow$ corresponding bounding-box GT annotation of the defect instance location.
    \end{itemize}
    
\end{enumerate}

\subsection{Training methodology}

All experiments were conducted on Nvidia A100 40GB GPUs. The baseline YOLO-NAS\cite{yolo-nas} model with three different architecture variants as  large, medium, and small are trained. For comparison analysis with previous work, we trained Super-YOLOv5\cite{zhang2023superyolo}. Additionally, we trained the proposed SEMI-SuperYOLO-NAS with integrated SR-branch, utilizing the default data augmentation strategy following the YOLO-NAS baseline training approach. Finally, we trained the proposed SEMI-SuperYOLO-NAS with integrated SR-branch using our suggested data augmentation strategy outlined in Section 3.2, replacing the previous default data augmentation strategy.

\section{Results and Discussion}
\begin{figure}[ht]
\centering

\begin{minipage}{0.19\textwidth}
\centering
\includegraphics[width=\textwidth]{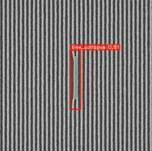}
\end{minipage}%
\hfill
\begin{minipage}{0.19\textwidth}
\centering
\includegraphics[width=\textwidth]{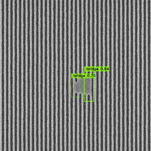}
\end{minipage}%
\hfill
\begin{minipage}{0.19\textwidth}
\centering
\includegraphics[width=\textwidth]{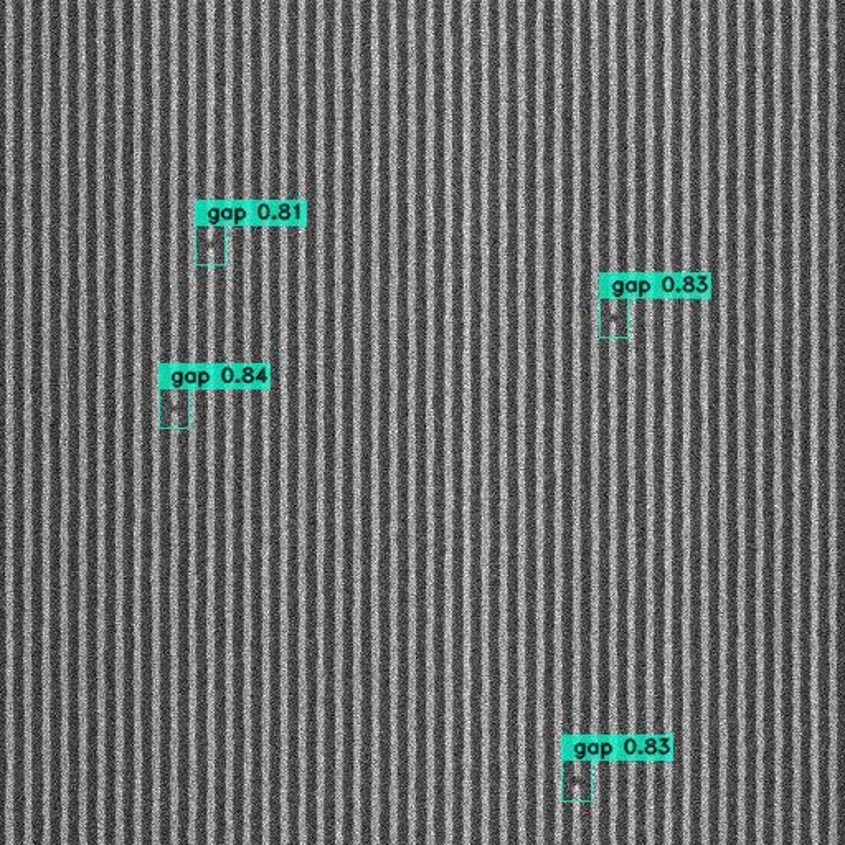}
\end{minipage}%
\hfill
\begin{minipage}{0.19\textwidth}
\centering
\includegraphics[width=\textwidth]{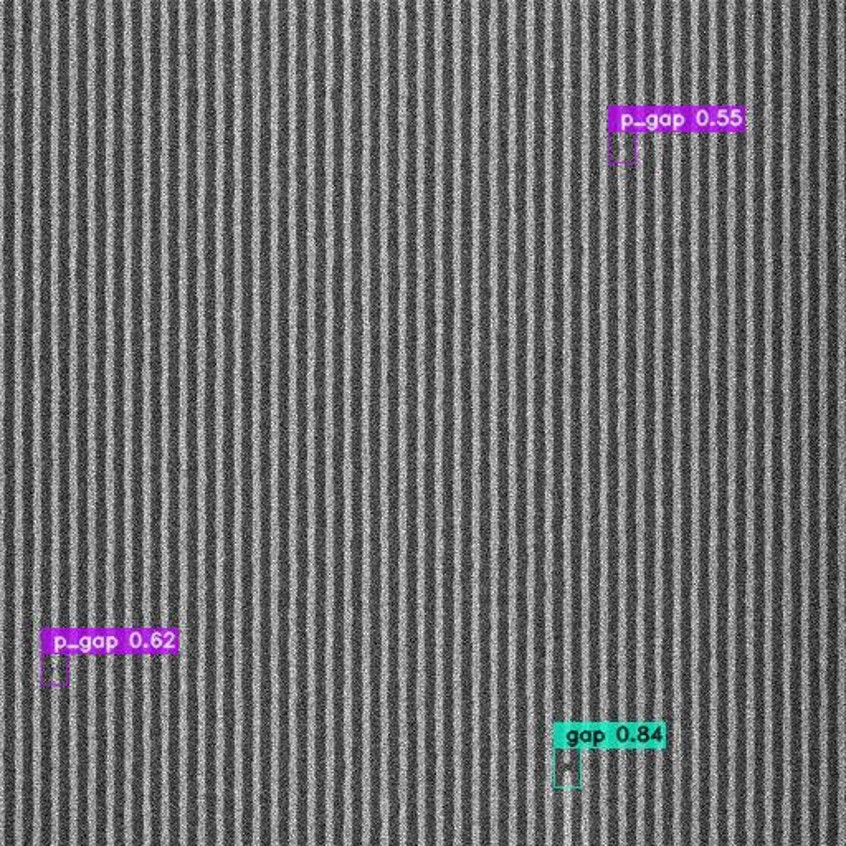}
\end{minipage}%
\hfill
\begin{minipage}{0.19\textwidth}
\centering
\includegraphics[width=\textwidth]{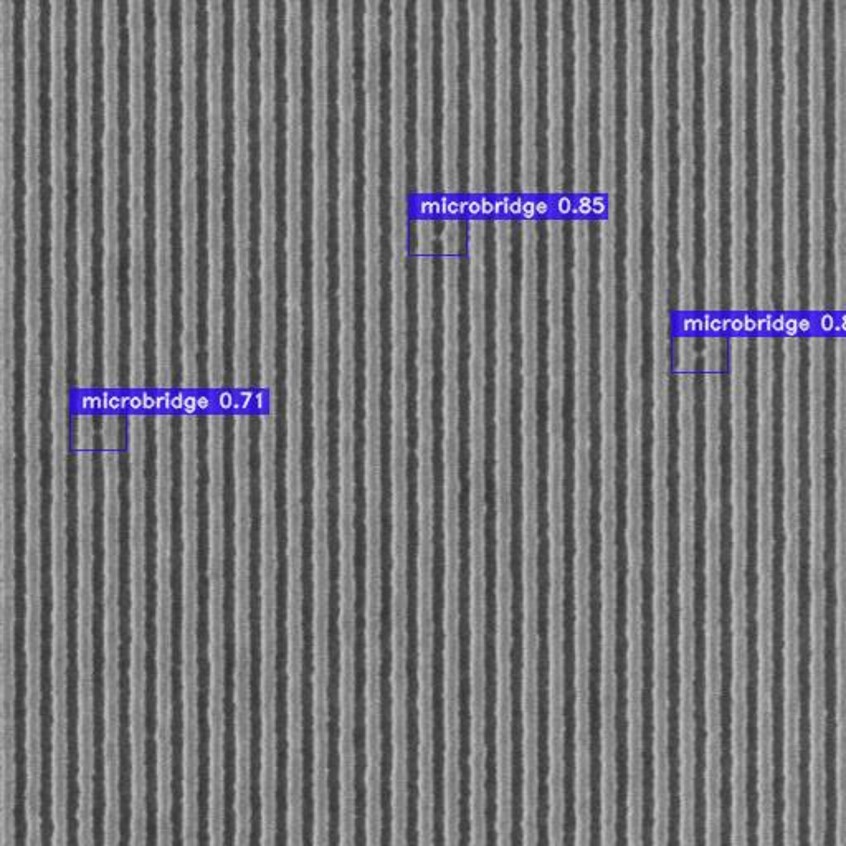}
\end{minipage}%
\hfill
\begin{minipage}{\linewidth}
\centering
\vspace{4pt}
(a)
\end{minipage}%

\vspace{6pt}

\begin{minipage}{0.19\linewidth}
\centering
\includegraphics[width=\textwidth]
{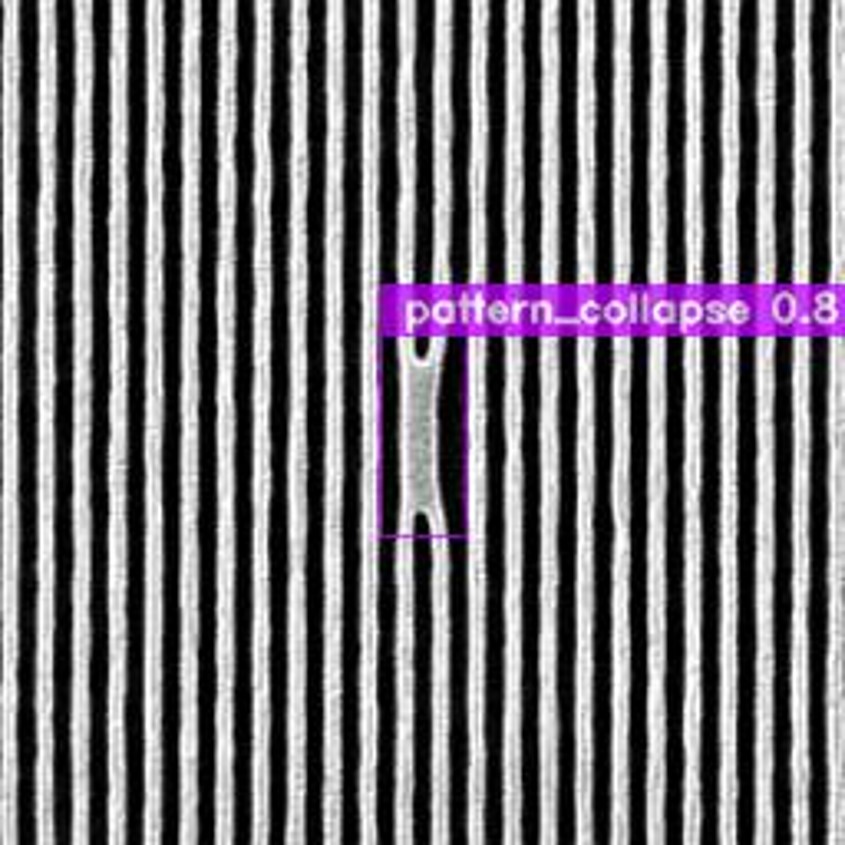}
\end{minipage}%
\hfill
\begin{minipage}{0.19\linewidth}
\centering
\includegraphics[width=\textwidth]{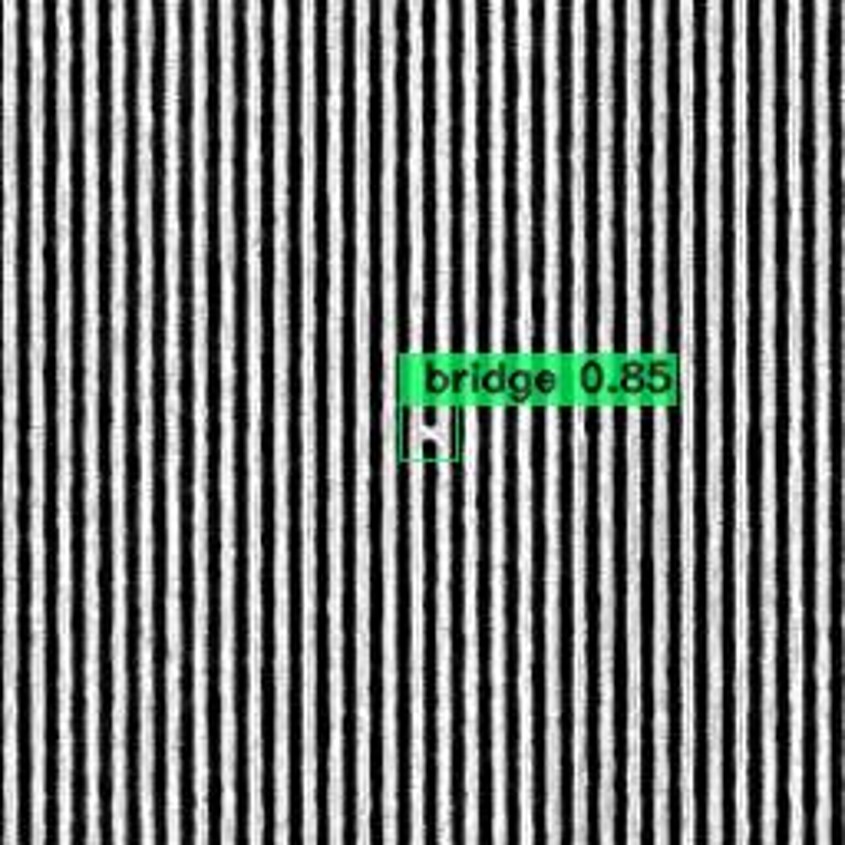}
\end{minipage}%
\hfill
\begin{minipage}{0.19\linewidth}
\centering
\includegraphics[width=\textwidth]{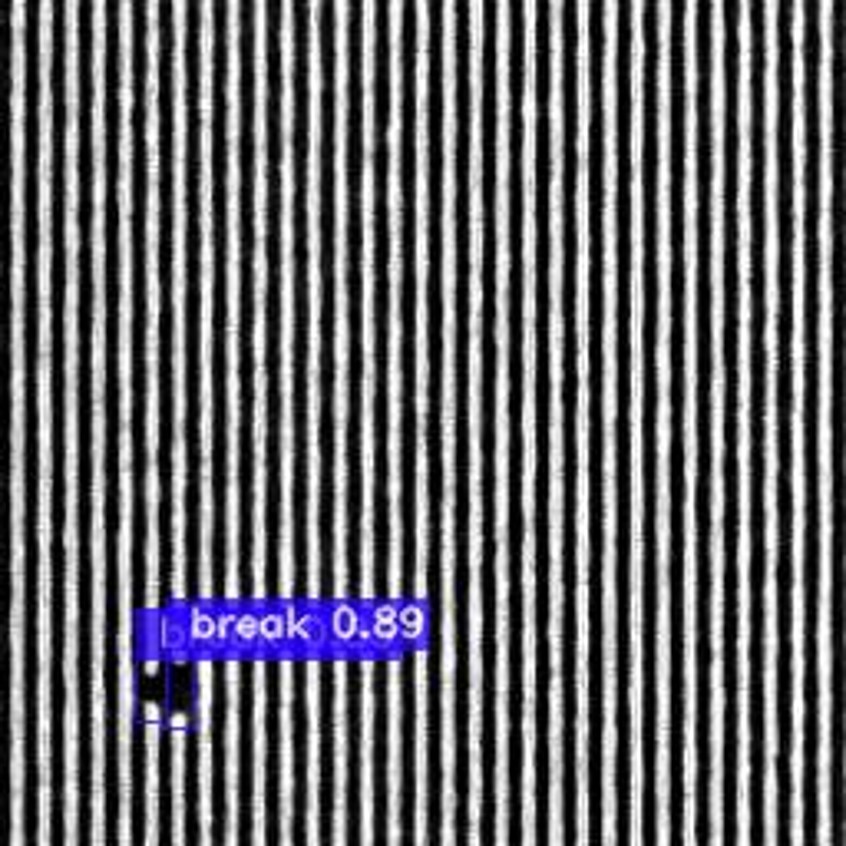}
\end{minipage}%
\hfill
\begin{minipage}{0.19\linewidth}
\centering
\includegraphics[width=\textwidth]{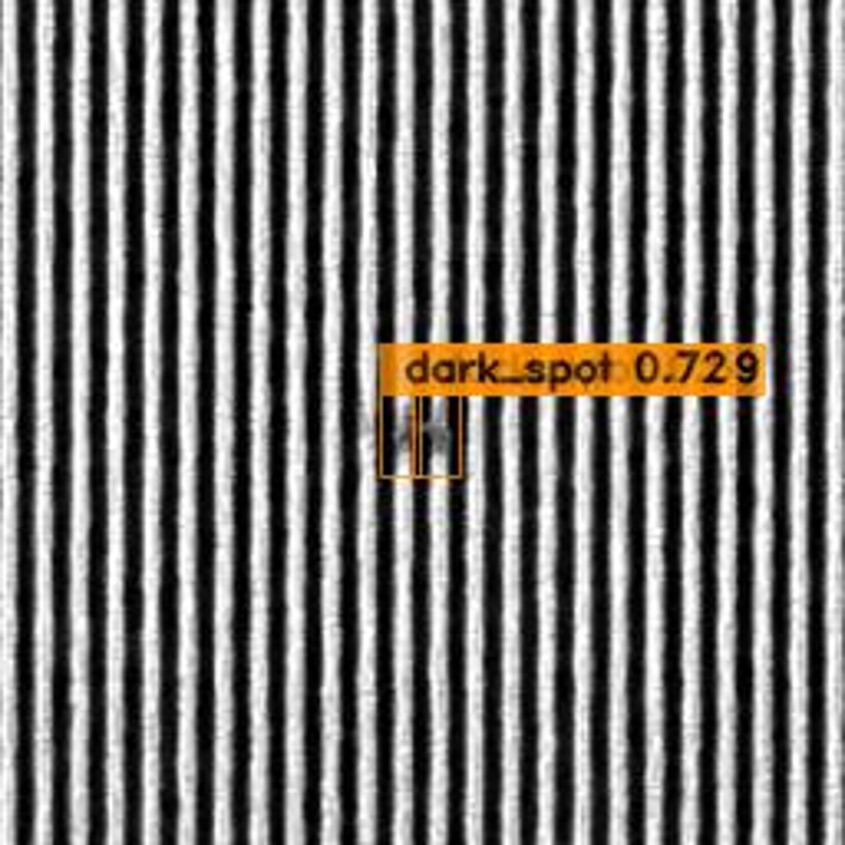}
\end{minipage}%

\begin{minipage}{\linewidth}
\centering
\vspace{4pt}
(b)
\end{minipage}%
\vspace{6pt}
\caption{Inference results on (a) SEM-ADI test set [image resolution pair (512 $\longrightarrow$ 1024)] (upper) and on (b) EDR-AEI test set [image resolution pair (256 $\longrightarrow$ 512)] (bottom) of our proposed SEMI-SuperYOLO-NAS architecture.}
\label{inference}
\end{figure}

The inference results of our proposed SEMI-SuperYOLO-NAS model are depicted in Fig.\ref{inference}(a)(b). Our discussion and analysis are divided into three sections:

\begin{enumerate}
    \item Firstly, we benchmarked our proposed approach (SEMI-SuperYOLO-NAS with our proposed data augmentation strategy) against baseline YOLO-NAS, previous SuperYOLOv5\cite{zhang2023superyolo}, and SEMI-SuperYOLO-NAS with the default data augmentation strategy. This benchmarking was conducted on the SEM-ADI dataset for image resolution pair (512 $\longrightarrow$ 1024) and on the EDR-AEI dataset for image resolution pair (240 $\longrightarrow$ 480).
    \item We demonstrate zero-shot inference on a new CD-SEM test dataset, comprising image pairs captured at the exact same locations but with two different resolutions (1024$\times$1024 and 512$\times$512). This newly collected dataset, also a SEM-ADI dataset, originates from a process condition distinct from the training dataset and possesses different CD/Pitch characteristics.The observable defects in these images closely resemble line-collapse and bridge/micro-bridge defects observed in the training images, albeit not identical.
    \item Finally, we assess the benefit of employing the SR-branch of our proposed SEMI-SuperYOLO-NAS model.  This involves generating or reconstructing upscaled images from their corresponding downscaled counterparts (e.g., 512 $\longrightarrow$ 1024) and conducting detection inference by applying baseline YOLO-NAS on these artificially upscaled images. This approach aims to facilitate more precise nano-scale defect inspection across various image resolutions without the need for explicit training. 
\end{enumerate}

\subsection{Benchmarking proposed SEMI-SuperYOLO-NAS ADCD framework}

We assessed four models using different strategies on the SEM-ADI dataset for the resolution pair (LR/HR) [512$\longrightarrow$ 1024] and on the EDR-AEI dataset for the resolution pair (LR/HR) [240 $\longrightarrow$ 480], and provided a comparative analysis for:

\begin{enumerate}
    \item YOLO-NAS (baseline)\cite{yolo-nas}, 
    \item Super-YOLOv5\cite{zhang2023superyolo},
    \item Proposed SEMI-SuperYOLO-NAS with default YOLO-NAS data augmentation strategy,
    \item Proposed SEMI-SuperYOLO-NAS with proposed data augmentation strategy outlined in Section 3.2
\end{enumerate}

Table \ref{ADI_result} illustrates that SuperYOLOv5\cite{zhang2023superyolo} outperforms all other defect detector models on the SEM-ADI dataset. However, as we propose a novel ADCD framework based on the recent YOLO-NAS architecture, our primary objective is to enhance its detection capability to match the performance of SuperYOLOv5\cite{zhang2023superyolo}. The baseline YOLO-NAS architecture performs the worst for all defect types, achieving a mAP of 0.47. While the performance for relatively prominent defect types like line-collapse and bridge is still comparable, it severely struggles with more challenging nano-scale defects such as gap, probable-gap, and micro-bridge. It is possible that the modules of the baseline YOLO-NAS architecture, such as the backbone, neck, and context modules, are not equipped to handle the intricacies of nano-scale defect detection, thus failing to meet the requirements in these aggressive pitches.

\begin{figure}[!ht]
\centering
\begin{minipage}{0.24\textwidth}
\centering
\includegraphics[width=\textwidth]{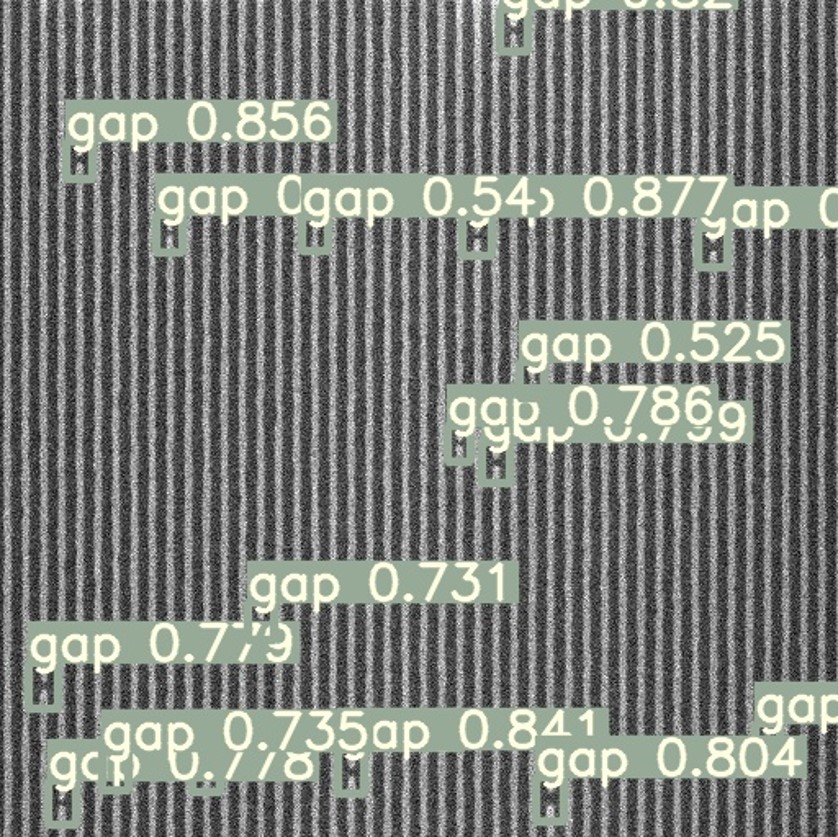}
(a)
\end{minipage}%
\hfill
\begin{minipage}{0.24\textwidth}
\centering
\includegraphics[width=\textwidth]{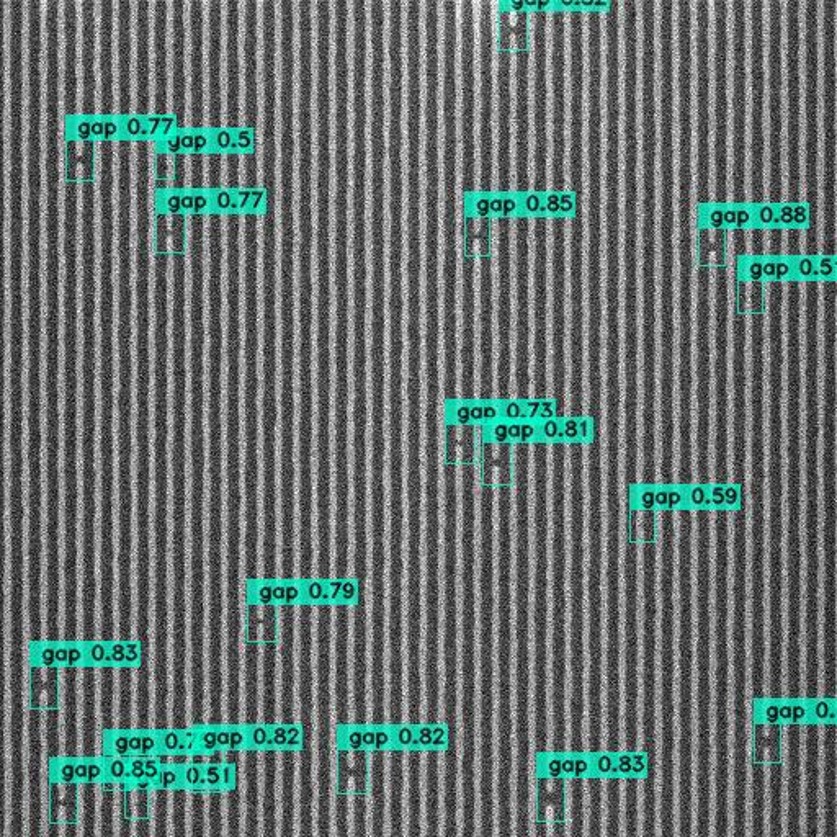}
(b)
\end{minipage}%
\hfill
\begin{minipage}{0.24\textwidth}
\centering
\includegraphics[width=\textwidth]{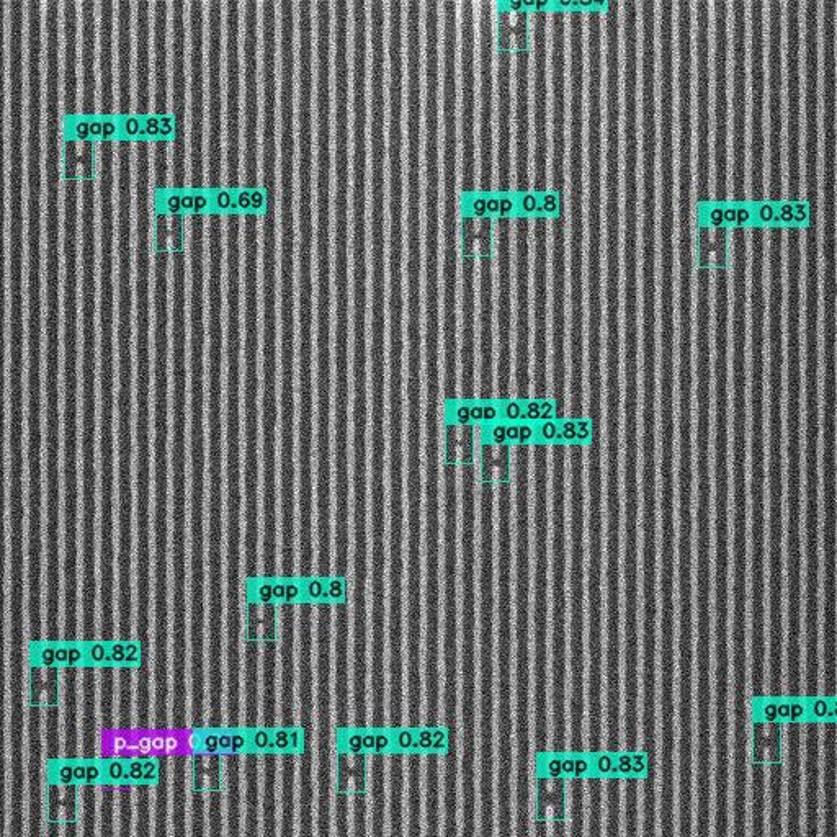}
(c)
\end{minipage}%
\vspace{6pt}
\caption{Detection results of gap and p-gap: (a) Baseline YOLO-NAS (b) Proposed SEMI-SuperYOLO-NAS + Default data augmentation (c) Proposed SEMI-SuperYOLO-NAS + proposed data augmentation. **False-Positive detections have been removed/reduced with our proposed data augmentation strategy towards improved detection accuracy}
\label{p-gap}
\end{figure}

\begin{figure}[!ht]
\centering
\begin{minipage}{0.24\textwidth}
\centering
\includegraphics[width=\textwidth]{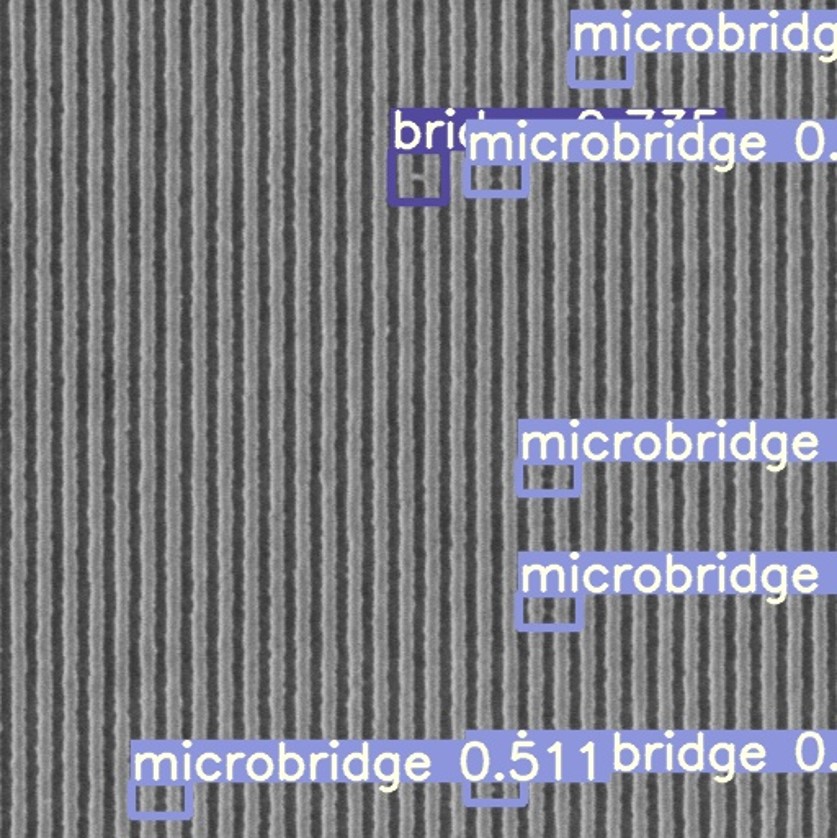}
(a)
\end{minipage}%
\hfill
\begin{minipage}{0.24\textwidth}
\centering
\includegraphics[width=\textwidth]{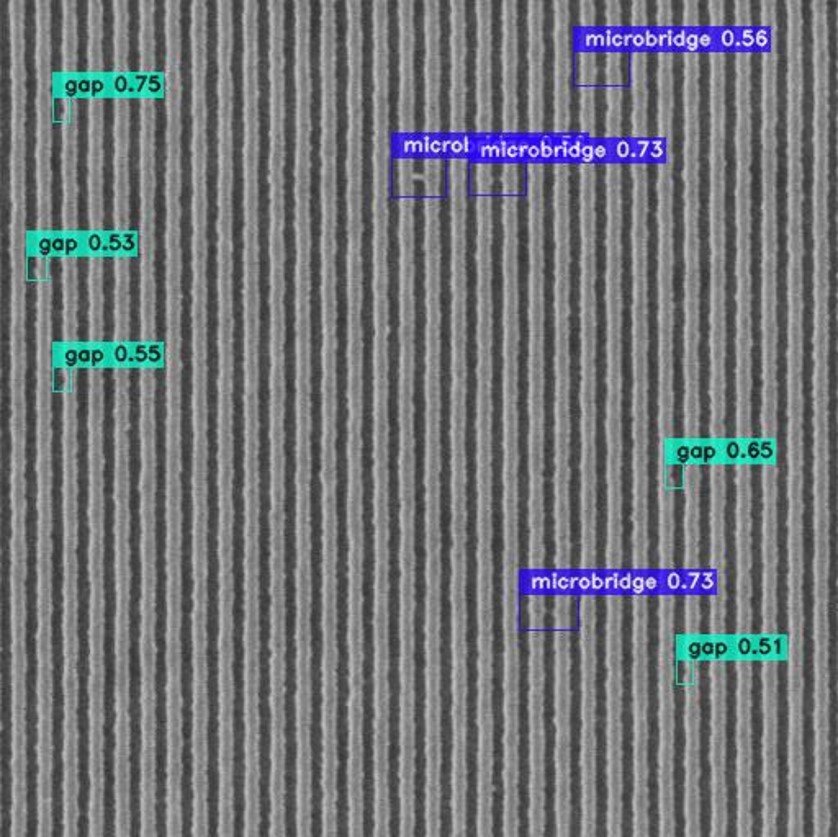}
(b)
\end{minipage}%
\hfill
\begin{minipage}{0.24\textwidth}
\centering
\includegraphics[width=\textwidth]{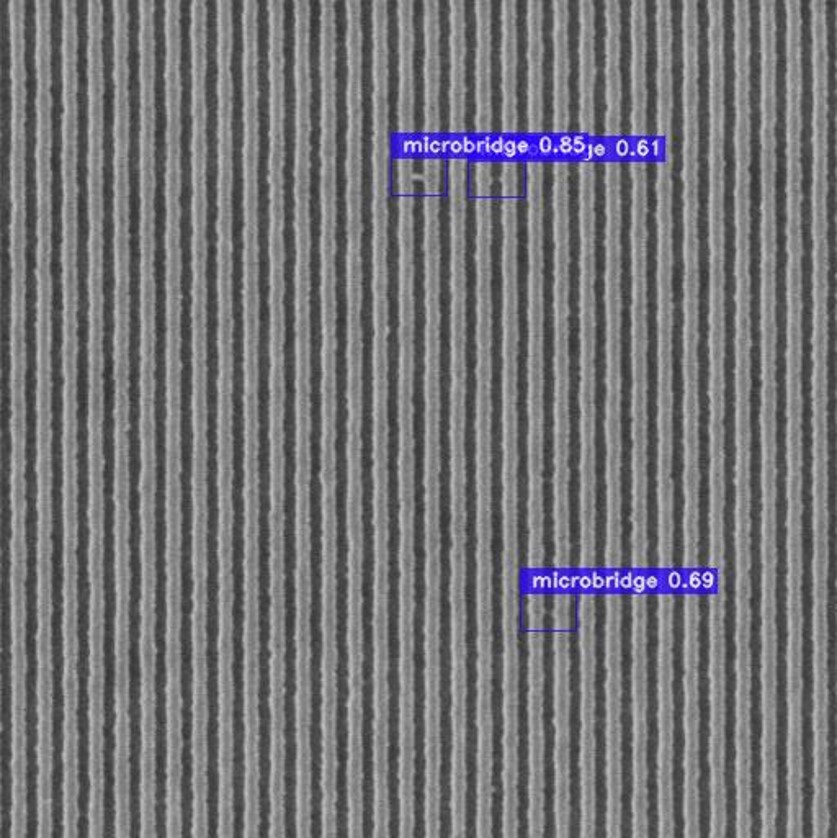}
(c)
\end{minipage}%
\vspace{6pt}
\caption{Detection results of micro-bridge: (a) Baseline YOLO-NAS (b) Proposed SEMI-SuperYOLO-NAS + Default data augmentation (c) Proposed SEMI-SuperYOLO-NAS + proposed data augmentation. **False-Positive detections have been removed/reduced with our proposed data augmentation strategy towards improved detection accuracy. }
\label{microbridge}
\end{figure}

In contrast, our proposed SEMI-SuperYOLO-NAS outperforms the baseline YOLO-NAS\cite{yolo-nas} architecture in both per class AP and mAP metrics, achieving an mAP of 0.54 with the default YOLO-NAS data augmentation strategy and a slightly improved mAP of 0.58 with the proposed data augmentation strategy, which replaces the previous default strategy. For micro-bridge defects, the AP improves by 51.06\% and 57.44\% with the corresponding data augmentation strategies when applying the proposed SEMI-SuperYOLO-NAS architecture compared to the baseline YOLO-NAS. The probable gap defect was not detected at all by the baseline YOLO-NAS model, whereas the proposed SEMI-SuperYOLO-NAS architecture, with both data augmentation strategies, achieved a per class AP of up to 0.13. For challenging gap defects, the proposed SEMI-SuperYOLO-NAS with the proposed augmentation strategy improves the AP from 0.15 to 0.19 ($\sim$27\%). Fig.\ref{p-gap} (c) and Fig.\ref{microbridge}(c) demonstrate improved detection results on these challenging nano-scale defects using our proposed approach (SEMI-SuperYOLO-NAS with the proposed data augmentation strategy), while also reducing significant false-positive detections. Therefore, integrating the SR-assisted branch definitely facilitates high-resolution feature learning by the defect detection backbone across all image resolutions, thus aiding the proposed approach to outperform its corresponding baseline model. Fig.\ref{ADI_AP} represents the metrics in Table \ref{ADI_result}.

\begin{figure}[!ht]
\centering
\begin{minipage}{0.24\textwidth}
\centering
\includegraphics[width=\textwidth]{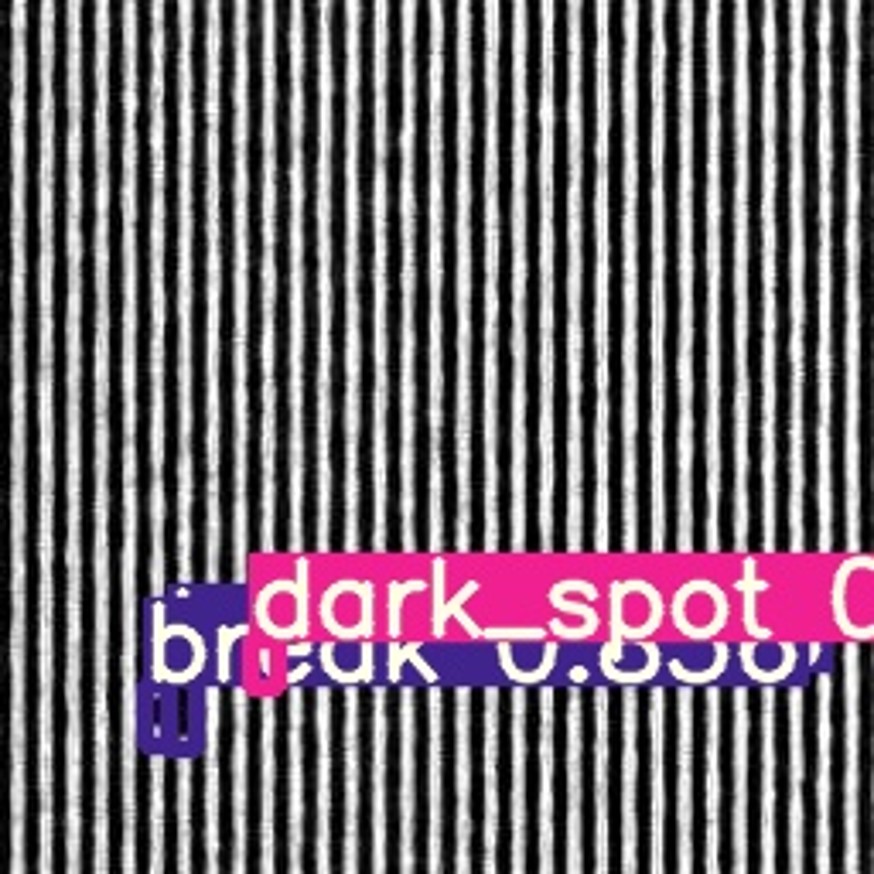}
(a)
\end{minipage}%
\hfill
\begin{minipage}{0.24\textwidth}
\centering
\includegraphics[width=\textwidth]{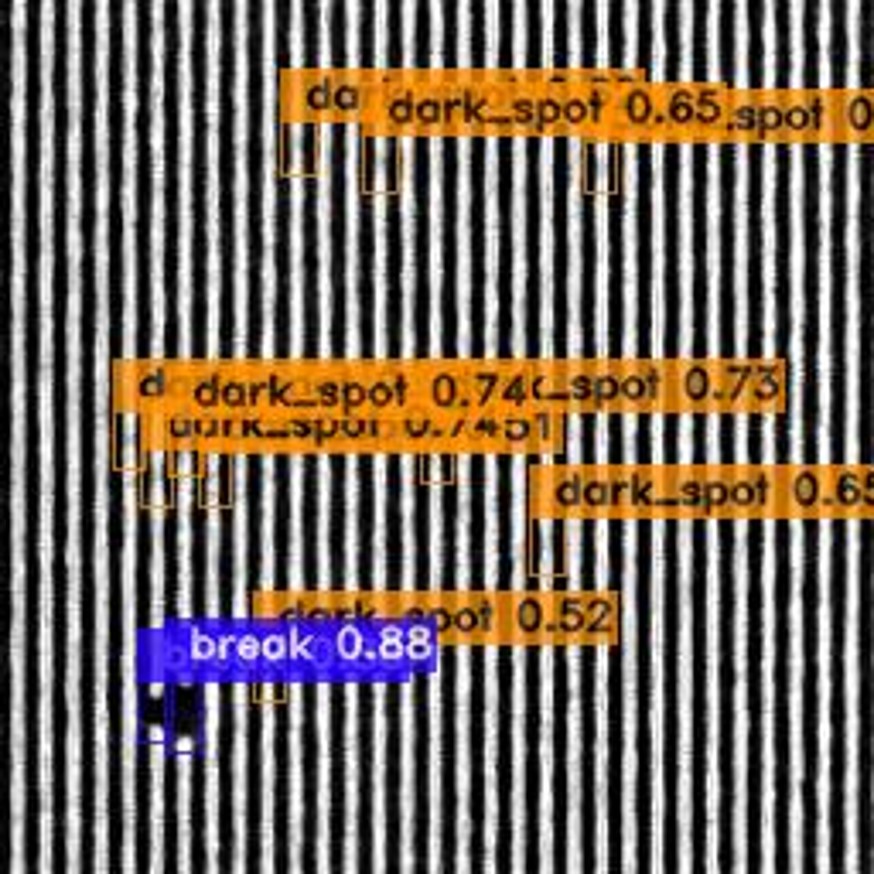}
(b)
\end{minipage}%
\hfill
\begin{minipage}{0.24\textwidth}
\centering
\includegraphics[width=\textwidth]{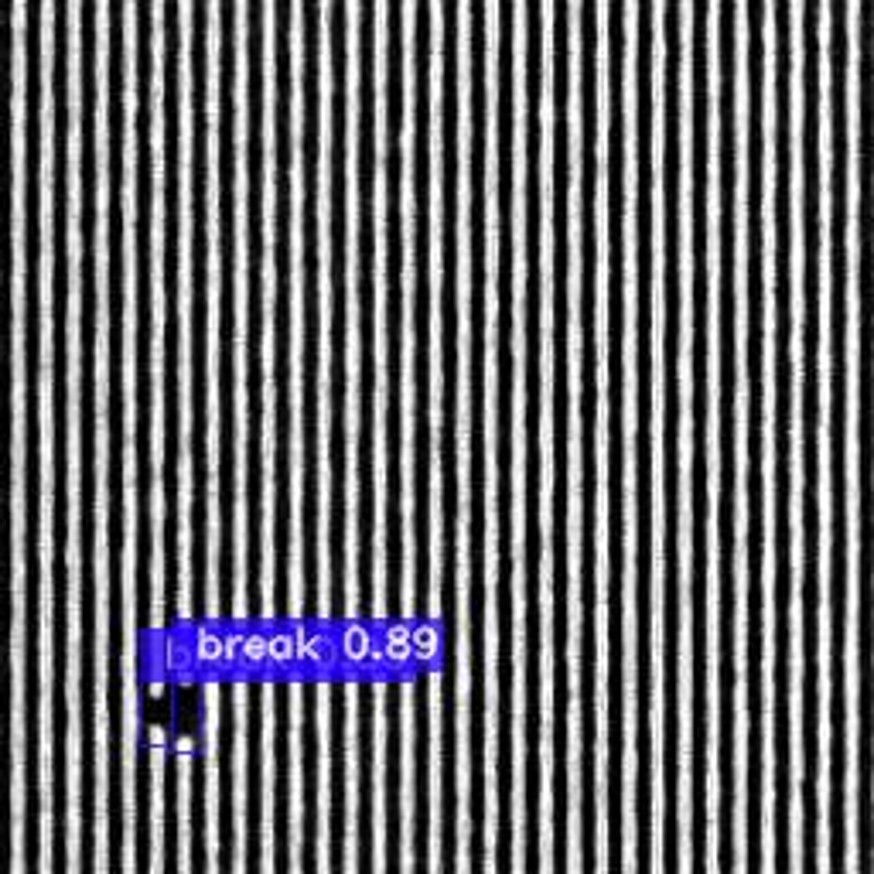}
(c)
\end{minipage}%
\vspace{6pt}
\caption{Detection results of break: (a) Baseline YOLO-NAS (b) Proposed SEMI-SuperYOLO-NAS + Default data augmentation (c) Proposed SEMI-SuperYOLO-NAS + proposed data augmentation. **False-Positive detections have been removed/reduced with our proposed data augmentation strategy towards improved detection accuracy.}
\label{break}
\end{figure}

\begin{figure}[!ht]
\vspace{10pt}
\centering
\begin{minipage}{0.24\textwidth}
\centering
\includegraphics[width=\textwidth]{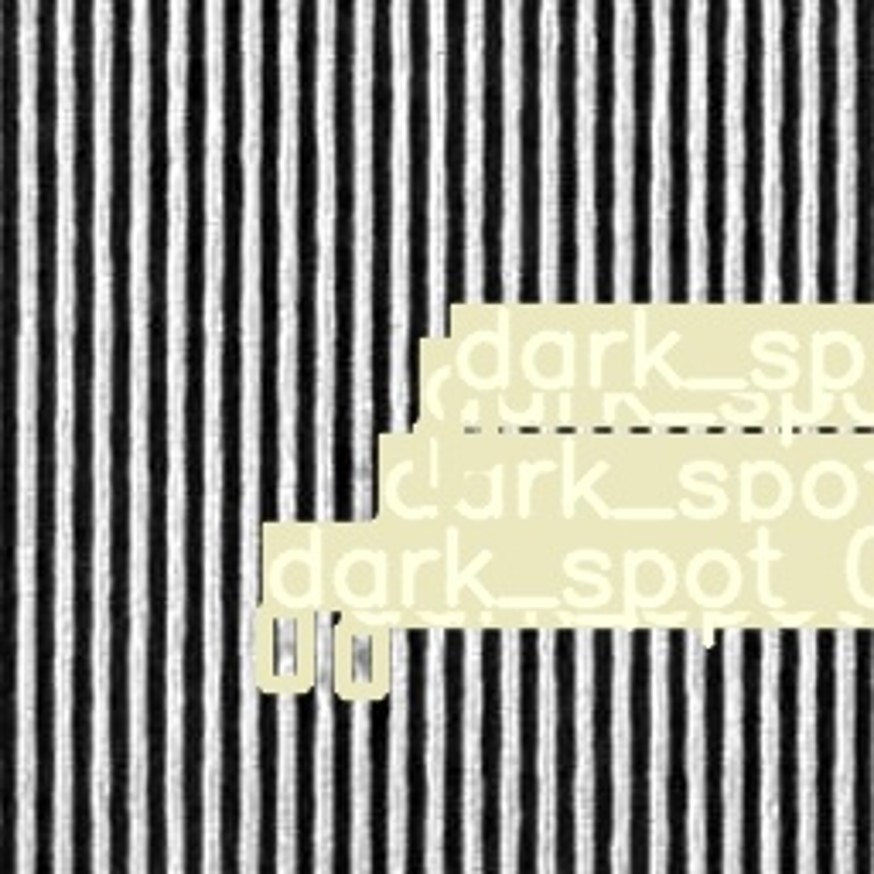}
(a)
\end{minipage}%
\hfill
\begin{minipage}{0.24\textwidth}
\centering
\includegraphics[width=\textwidth]{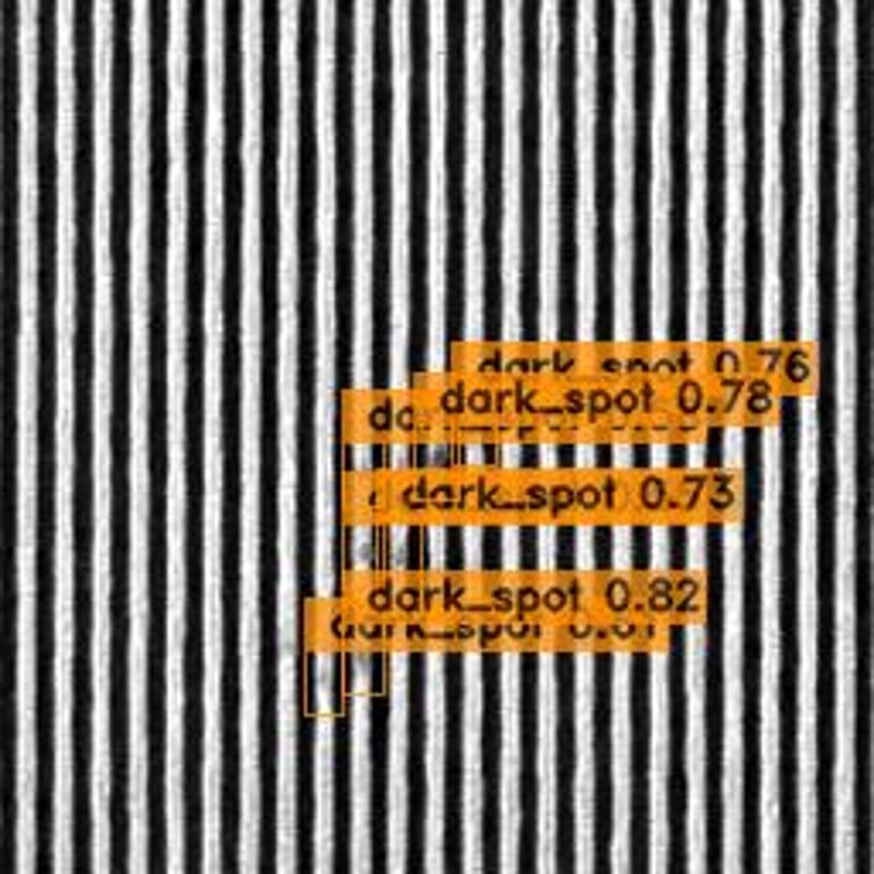}
(b)
\end{minipage}%
\hfill
\begin{minipage}{0.24\textwidth}
\centering
\includegraphics[width=\textwidth]{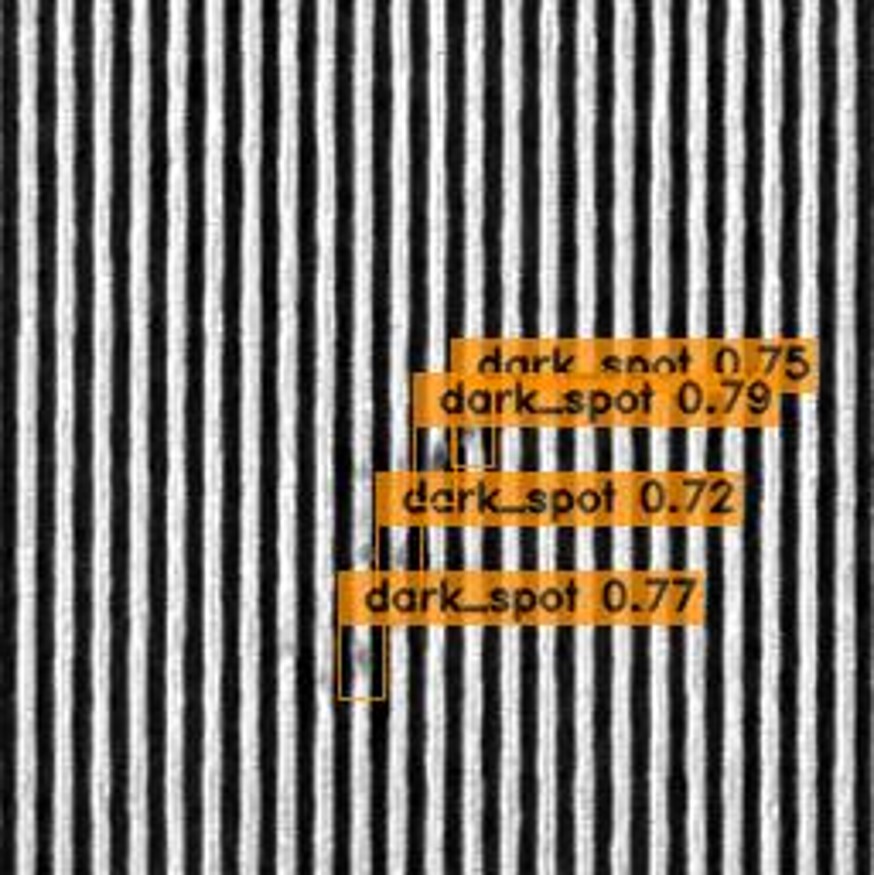}
(c)
\end{minipage}%
\vspace{6pt}
\caption{Detection results of dark-spot: (a) Baseline YOLO-NAS (b) Proposed SEMI-SuperYOLO-NAS + Default data augmentation (c) Proposed SEMI-SuperYOLO-NAS + proposed data augmentation. **False-Positive detections have been removed/reduced with our proposed data augmentation strategy towards improved detection accuracy.}
\label{darkspot}
\end{figure}

\begin{figure}[ht]
    \centering
    
    \begin{minipage}{0.75\textwidth}
        \includegraphics[width=\linewidth]{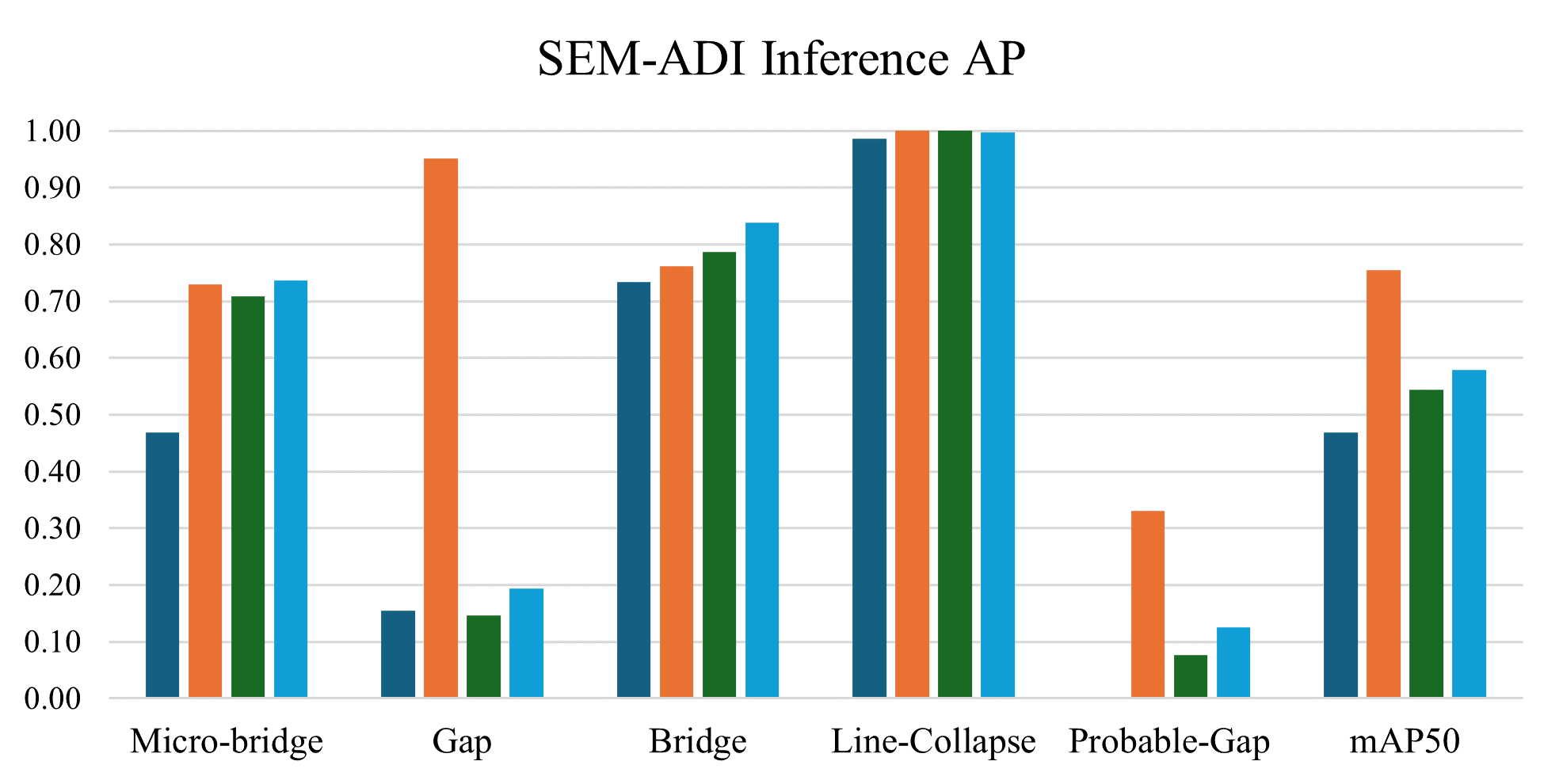}
        
    \end{minipage}%
    \hfill
    \begin{minipage}{0.25\textwidth}        
            \smallskip
            \centering
            \begin{tabular}{c l}
                
                \vspace{6pt}
    
                \begin{tikzpicture}
                    \fill[custom_dark_blue] (0,0) rectangle (0.25,0.25);
                \end{tikzpicture} & \footnotesize \begin{tabular}[t]{@{}l@{}} YOLO-NAS \\ (baseline) \\ \end{tabular} \\
                \vspace{6pt}
                
                \begin{tikzpicture}
                \small
                    \fill[custom_orange] (0,0) rectangle (0.25,0.25);
                \end{tikzpicture} & \footnotesize \begin{tabular}[t]{@{}l@{}} *SuperYOLOv5\cite{zhang2023superyolo} \\ \end{tabular} \\
                \vspace{6pt}
                
                \begin{tikzpicture}
                    \fill[custom_dark_green] (0,0) rectangle (0.25,0.25);
                \end{tikzpicture} & \footnotesize \begin{tabular}[t]{@{}l@{}} Proposed \\SEMI-SuperYOLO-NAS \\ (Default Aug) \\ \end{tabular} \\
                \vspace{6pt}
                
                \begin{tikzpicture}
                    \fill[custom_light_blue] (0,0) rectangle (0.25,0.25);
                \end{tikzpicture} & \footnotesize \begin{tabular}[t]{@{}l@{}} Proposed \\ SEMI-SuperYOLO-NAS \\ (Proposed Aug) \\ \end{tabular} \\
                \vspace{6pt}
    
            \end{tabular}
    
        \end{minipage}
        
    \vspace{6pt}
    \caption{Comparison analysis of defect detection performance [per class AP and mean AP] on SEM-ADI dataset. [**The original SEM image of resolution 1024 was downsampled to 512, and all detector architecture variants are Small].}
    \label{ADI_AP}
\end{figure}

\begin{figure}[!ht]
\vspace{6pt}

\centering
\begin{minipage}{0.75\textwidth}
    \includegraphics[width=\linewidth]{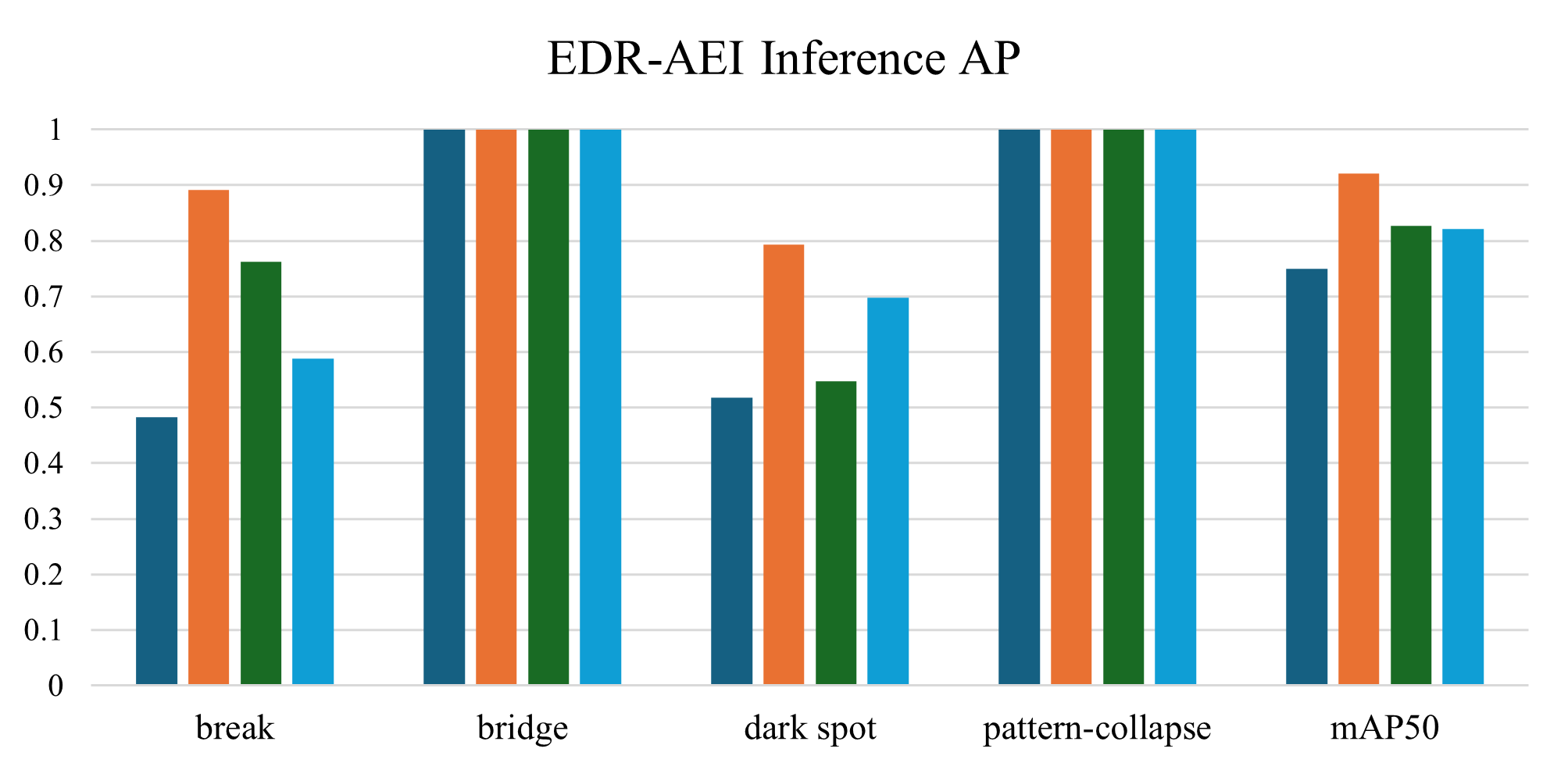}
    
\end{minipage}%
\hfill
\begin{minipage}{0.25\textwidth}        
        \smallskip
        \centering
        \begin{tabular}{c l}
            \vspace{6pt}

            \begin{tikzpicture}
                \fill[custom_dark_blue] (0,0) rectangle (0.25,0.25);
            \end{tikzpicture} & \footnotesize \begin{tabular}[t]{@{}l@{}} YOLO-NAS \\ (baseline) \\ \end{tabular} \\
            \vspace{6pt}
            
            \begin{tikzpicture}
            \small
                \fill[custom_orange] (0,0) rectangle (0.25,0.25);
            \end{tikzpicture} & \footnotesize \begin{tabular}[t]{@{}l@{}} *SuperYOLOv5\cite{zhang2023superyolo} \\ \end{tabular} \\
            \vspace{6pt}
            
            \begin{tikzpicture}
                \fill[custom_dark_green] (0,0) rectangle (0.25,0.25);
            \end{tikzpicture} & \footnotesize \begin{tabular}[t]{@{}l@{}} Proposed \\SEMI-SuperYOLO-NAS \\ (Default Aug)\\ \end{tabular} \\
            \vspace{6pt}
            
            \begin{tikzpicture}
                \fill[custom_light_blue] (0,0) rectangle (0.25,0.25);
            \end{tikzpicture} & \footnotesize \begin{tabular}[t]{@{}l@{}} Proposed \\ SEMI-SuperYOLO-NAS \\ (Proposed Aug) \\ \end{tabular} \\
            \vspace{6pt}

        \end{tabular}

    \end{minipage}
\vspace{6pt}
\caption{Comparison analysis of defect detection performance [per class AP and mean AP] on EDR-AEI dataset. [**The original SEM image of resolution 512 was downsampled to 256, and all detector architecture variants are Small].}
\label{EDR_AP}
\end{figure}

Similar trends are observed in the EDR-AEI dataset, as shown in Table \ref{EDR_result} and depicted in Fig.\ref{EDR_AP}. Compared to the baseline YOLO-NAS (with a mAP of 0.75), our proposed SEMI-SuperYOLO-NAS architecture, incorporating both data augmentation strategies, achieved a mAP of approximately 0.83, marking a 10.67\% improvement. While the performance of all four models remains consistent for pattern-collapse and bridge defects, an intriguing trade-off is evident for dark-spot and break defects between the proposed SR-assisted branch and the proposed data augmentation strategy within the SEMI-SuperYOLO-NAS framework. Notably, the SEMI-SuperYOLO-NAS architecture, with both data augmentation strategies, outperforms the baseline YOLO-NAS, with the SR-assisted branch significantly enhancing the detection of break defect instances (increasing AP from 0.48 to 0.76) and the proposed data augmentation strategy markedly improving the detection of dark-spot instances (increasing AP from 0.52 to 0.70). Fig.\ref{break}(c) and Fig.\ref{darkspot}(c) demonstrate improved detection results on these challenging nano-scale defects using our proposed approach (SEMI-SuperYOLO-NAS with the proposed data augmentation strategy), while also reducing significant false-positive detections. This trade-off presents an intriguing avenue for future investigation to achieve a balanced approach

\begin{table}[htbp]

    \caption{Per class AP and mean AP @IoU 0.5 on SEM-ADI dataset. Best, Second best and Third best colour codes: \textcolor{green}{GREEN}, \textcolor{red}{RED}, \textcolor{blue}{BLUE}.}
    \centering
    \begin{tabular}{|c|c|c|c|c|}
        \hline
         & \multicolumn{4}{|c|}{\textbf{Defect Detector Models}} \\
        \hline
        Defect Class & \textbf{YOLO-NAS} & \textbf{SuperYOLOv5\cite{zhang2023superyolo}} & \textbf{Proposed SEMI-} & \textbf{Proposed SEMI-} \\
        &  \textbf{(baseline)} && \textbf{SuperYOLO-NAS} & \textbf{SuperYOLO-NAS} \\
        & & & \textbf{(+Default Aug)} & \textbf{(+Proposed Aug)} \\
        \hline
        \textbf{micro-bridge} & 0.47 & \textcolor{red}{0.73} & \textcolor{blue}{0.71} & \textcolor{green}{0.74} \\
        \hline
        \textbf{gap} & 0.15 & \textcolor{green}{0.95} & \textcolor{blue}{0.15} & \textcolor{red}{0.19} \\
        \hline
        \textbf{bridge} & 0.73 & \textcolor{blue}{0.76} & \textcolor{red}{0.79} & \textcolor{green}{0.84} \\
        \hline
        \textbf{line-collapse} & \textcolor{red}{0.99} & \textcolor{green}{1.00} & \textcolor{green}{1.00} & \textcolor{green}{1.00} \\
        \hline
        \textbf{probable gap} & 0.00 & \textcolor{green}{0.33} & \textcolor{blue}{0.08} & \textcolor{red}{0.13} \\
        \hline
        \textbf{mAP} & 0.47 & \textcolor{green}{0.75} & \textcolor{blue}{0.54} & \textcolor{red}{0.58} \\
        \hline
    \end{tabular}

    \label{ADI_result}
\end{table}

\begin{table}[htbp]
\vspace{15pt}
    \caption{Per class AP and mean AP @IoU 0.5 on EDR-AEI dataset. . Best, Second best and Third best colour codes: \textcolor{green}{GREEN}, \textcolor{red}{RED}, \textcolor{blue}{BLUE}.}
    \centering
     \begin{tabular}{|c|c|c|c|c|}
        \hline
         & \multicolumn{4}{|c|}{\textbf{Defect Detector Models}} \\
        \hline
        Defect Class & \textbf{YOLO-NAS} & \textbf{SuperYOLOv5\cite{zhang2023superyolo}} & \textbf{Proposed SEMI-} & \textbf{Proposed SEMI-} \\
        &  \textbf{(baseline)} && \textbf{SuperYOLO-NAS} & \textbf{SuperYOLO-NAS} \\
        & & & \textbf{(+Default Aug)} & \textbf{(+Proposed Aug)} \\
        \hline
        \textbf{break} & 0.48 & \textcolor{green}{0.89} & \textcolor{red}{0.76} & \textcolor{blue}{0.59} \\
        \hline
        \textbf{bridge} & \textcolor{green}{1.00} & \textcolor{green}{1.00} & \textcolor{green}{1.00} & \textcolor{green}{1.00} \\
        \hline
        \textbf{dark spot} & 0.52 & \textcolor{green}{0.79} & \textcolor{blue}{0.55} & \textcolor{red}{0.70} \\
        \hline
        \textbf{pattern coll.} & \textcolor{green}{1.00} & \textcolor{green}{1.00} & \textcolor{green}{1.00} & \textcolor{green}{1.00} \\
        \hline
        \textbf{mAP} & 0.75 & \textcolor{green}{0.92} & \textcolor{red}{0.83} & \textcolor{blue}{0.82} \\
        \hline
    \end{tabular}

    \label{EDR_result}
\end{table}

\subsection{Zero-shot inference with newly acquired High/Low resolution image pair (1024/512)}

\begin{figure}[ht]
\centering
\hfill

\begin{minipage}{0.60\textwidth}
\centering
\includegraphics[width=0.45\textwidth]{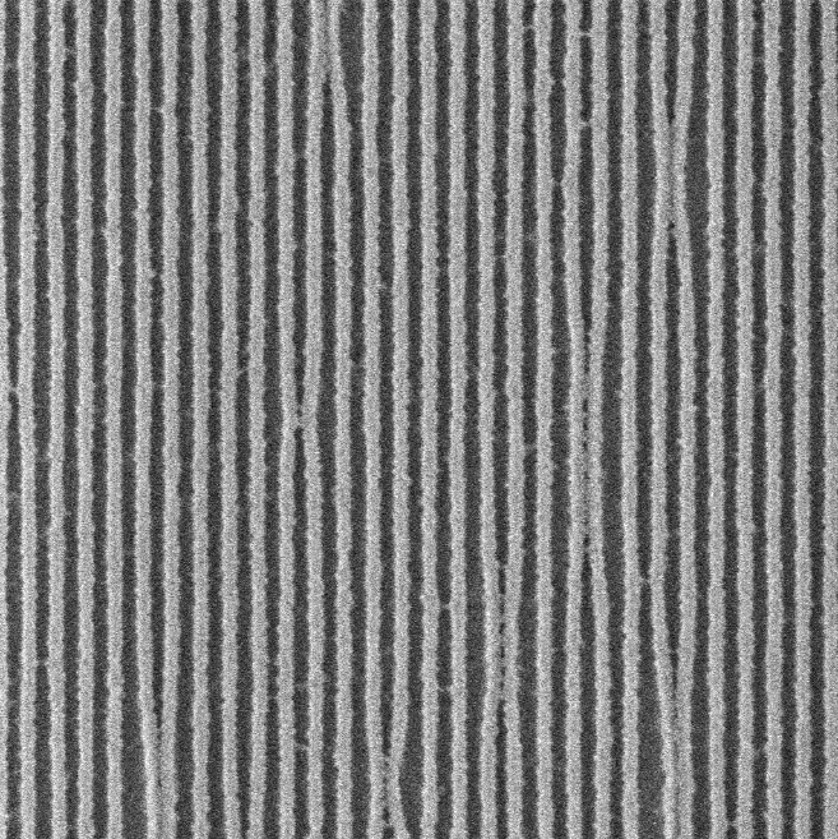}
\hfill
\includegraphics[width=0.45\textwidth]{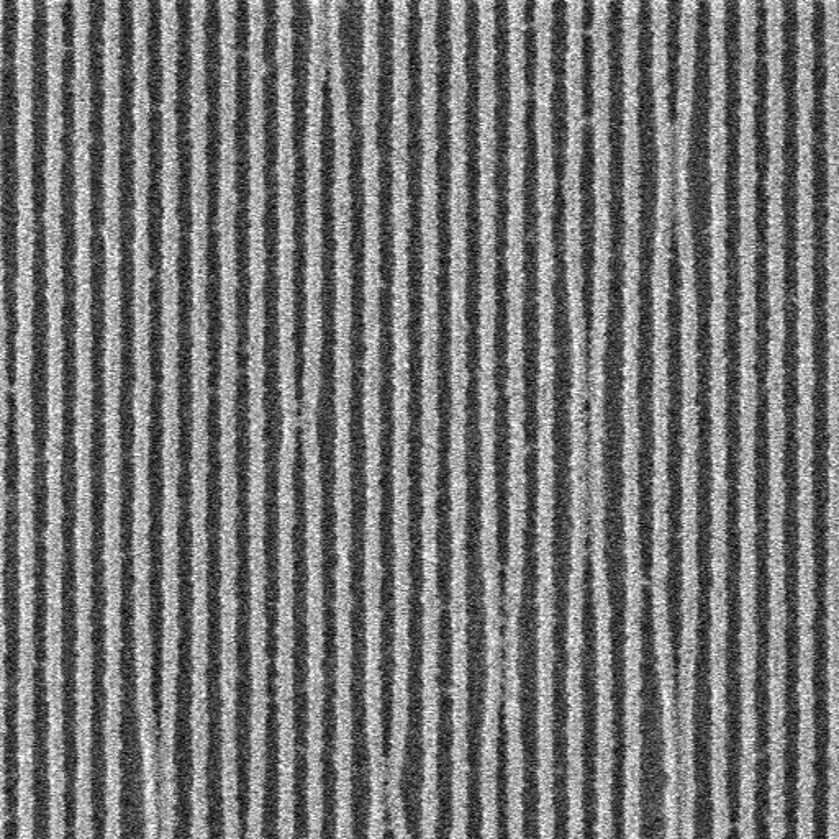}
\end{minipage}%
\\
\vspace{4pt}
\begin{minipage}{0.60\textwidth}
\centering

(a)
\hspace{0.475\textwidth}
(b)

\end{minipage}%
\caption{CD-SEM images of new collected defect at same location, (a) and (b) are 1024 and 512,respectively.}
\label{one_shot}

\end{figure}

\begin{figure}[!ht]
\vspace{10pt}
\centering
\includegraphics[width=8cm]{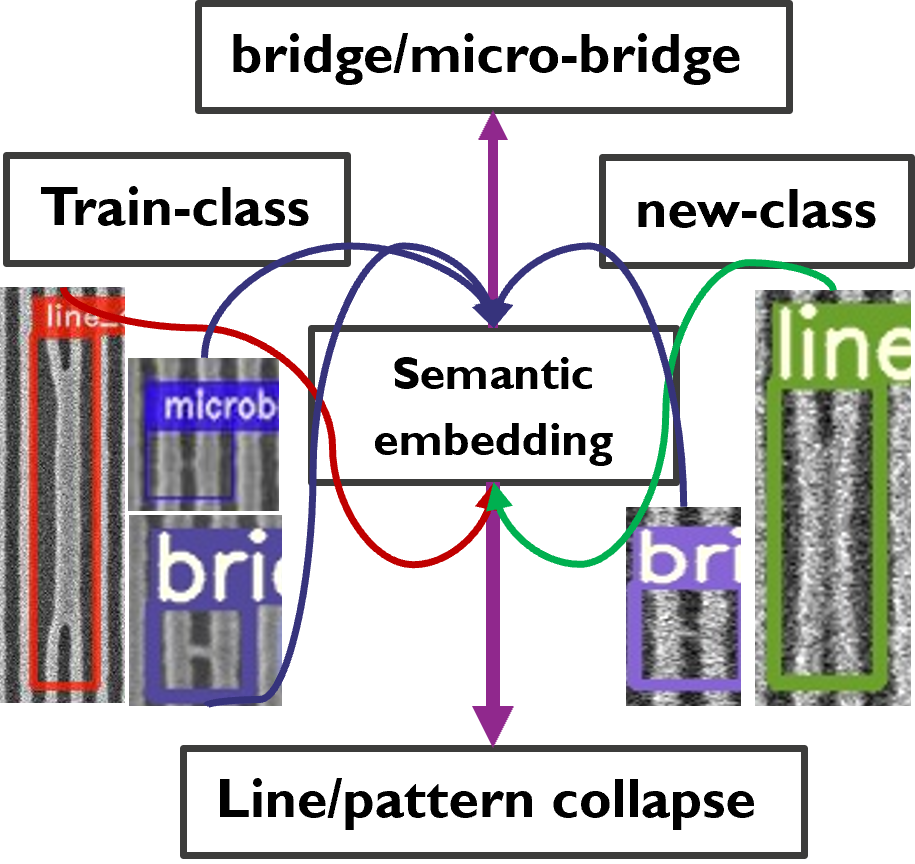}
\caption{Illustration of zero-shot inference: predictions on defect classes not encountered during training. A "weakly" semantic embedding can be observed between the defect "training" class (left) and the "inference" class (right) of two different process conditions}
\label{embeddings}
\vspace{10pt}
\end{figure}

\begin{figure}[!ht]
\centering
\includegraphics[width=5cm]{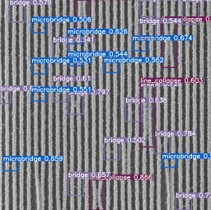}
\caption{Illustration of zero-shot inference: Detection of defect instances using baseline YOLO-NAS on new SEM-ADI test dataset (Line-Space) with resolution 1024 @confidence threshold 0.5. This dataset was collected from a process distinct from training dataset, with different CD/Pitch and imaging conditions. **YOLO-NAS model, trained with SEM-ADI dataset [as shown in Fig. 2] at a resolution of 1024, is employed for inference.}
\label{oneshot1024}
\vspace{10pt}
\end{figure}

\begin{figure}[!ht]
\centering
\begin{minipage}{0.24\textwidth}
\centering
\includegraphics[width=\textwidth]{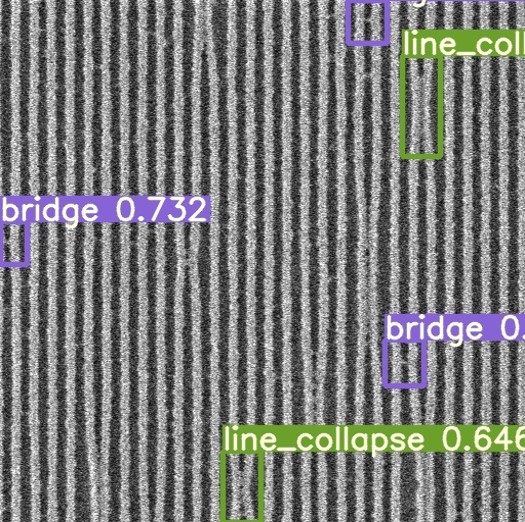}
(a)
\end{minipage}%
\hfill
\begin{minipage}{0.24\textwidth}
\centering
\includegraphics[width=\textwidth]{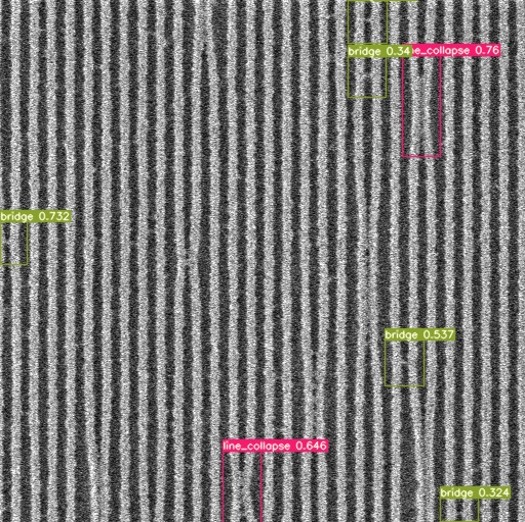}
(b)
\end{minipage}%
\hfill
\begin{minipage}{0.24\textwidth}
\centering
\includegraphics[width=\textwidth]{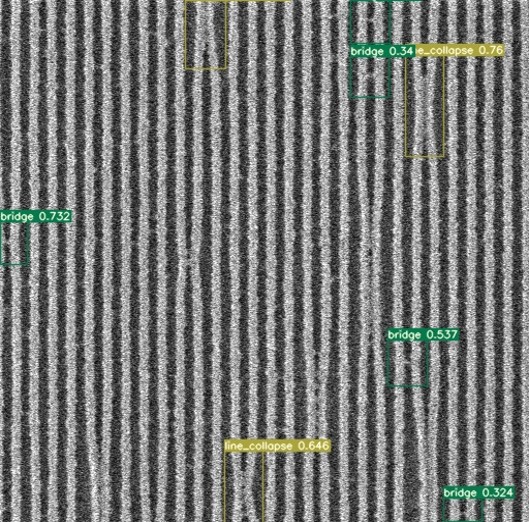}
(c)
\end{minipage}%
\hfill
\begin{minipage}{0.24\textwidth}
\centering
\includegraphics[width=\textwidth]{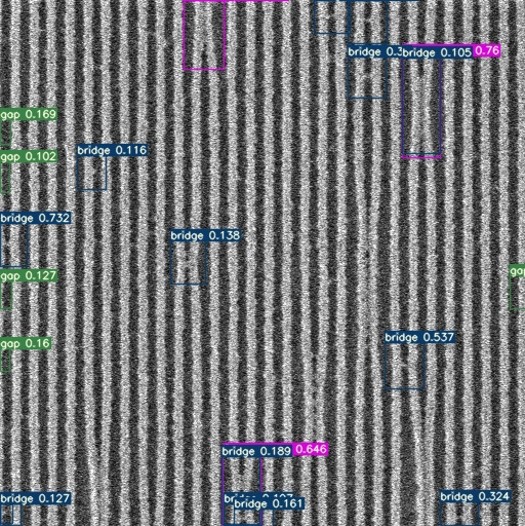}
(d)
\end{minipage}%
\vspace{6pt}

\caption{Illustration of zero-shot inference: Detection of defect instances using baseline YOLO-NAS on new SEM-ADI test dataset (Line-Space) with resolution 512 for different confidence thresholds. (a) 0.5, (b) 0.3, (c) 0.2, and (d)0.1. This dataset was collected from a process distinct from training dataset , with different CD/Pitch and imaging conditions.. **YOLO-NAS model, trained with the downsampled SEM-ADI dataset [as shown in Fig.\ref{defect}] at a resolution of 512, is employed for inference.}
\label{conf}
\vspace{10pt}
\end{figure}

To demonstrate zero-shot inference across different resolutions and real SEM imaging conditions, we collected a new SEM-ADI dataset from a process condition distinct from the training dataset, as depicted in Fig.\ref{one_shot}. This novel dataset has a different CD/Pitch and was obtained from the same location at two different resolutions: 512 and 1024, respectively, as shown in Fig.\ref{one_shot}. The observable defects in these images closely resemble line-collapse and bridge/micro-bridge defects seen in the training images. Our primary aim was to ascertain whether our proposed ADCD framework could obviate the need for explicit training across diverse image conditions (process, resolutions, CD/Pitch) and acquire "weakly" semantic embeddings between closely matched defect categories, as demonstrated in Fig.\ref{embeddings}. Fig.\ref{oneshot1024} depicts defect detection inference result (with confidence threshold $>=$ 0.5) on this newly acquired ADI-SEM image of resolution 1024, using the baseline YOLO-NAS model trained on the SEM-ADI dataset of the same resolution (dataset shown in Fig.\ref{one_shot}). Similarly, Fig.\ref{conf} illustrates the defect detection inference results (at different confidence thresholds: 0.1, 0.2, 0.3, and 0.5) on the identical dataset but at a lower resolution of 512. This was achieved using the same YOLO-NAS model trained on the previous SEM-ADI dataset of resolution 512 (dataset shown in Fig.\ref{one_shot}). These SEM images depicted in Fig.\ref{conf} are taken at the exact same location as Fig.\ref{oneshot1024} but with lower resolution using the imaging tool’s settings. Notable observations include a decrease in the detection of defect instances for the image resolution of 512 compared to 1024, and an increase in the detection of defect instances with a lower confidence threshold (for example, the model detected more instances with a confidence threshold of 0.1 compared to when the confidence threshold was 0.5). Lowering the confidence threshold is expected as this dataset was never seen by the model. This highlights an existing bottleneck with supervised machine learning applications that should be addressed in the near future.

\subsection{Inference with baseline YOLO-NAS on Super-Resolution images [Generated/Reconstructed with SR-assisted branch]}

The primary objective of this research is to propose a scale-variant ADCD framework capable of upscaling images to enable more accurate nano-scale defect inspection across various image resolutions without explicit training. In this experiment, we initially trained the baseline YOLO-NAS\cite{yolo-nas-deci} models on each image resolution, including 1024, 512, 256, 128, and 64, both with and without the proposed data augmentation strategy outlined in Section 3.2. We then reported the best mAP metric @IoU 0.5 for each resolution. The dataset preparation strategy for corresponding HR/LR image-pairs, as discussed in Section 3.3, was employed. We observed that the lowest resolution allowing for detection was 128$\times$128 with our experimental setup. Beyond this resolution, detecting defects became challenging due to the limited number of pixels defining the line and space.  

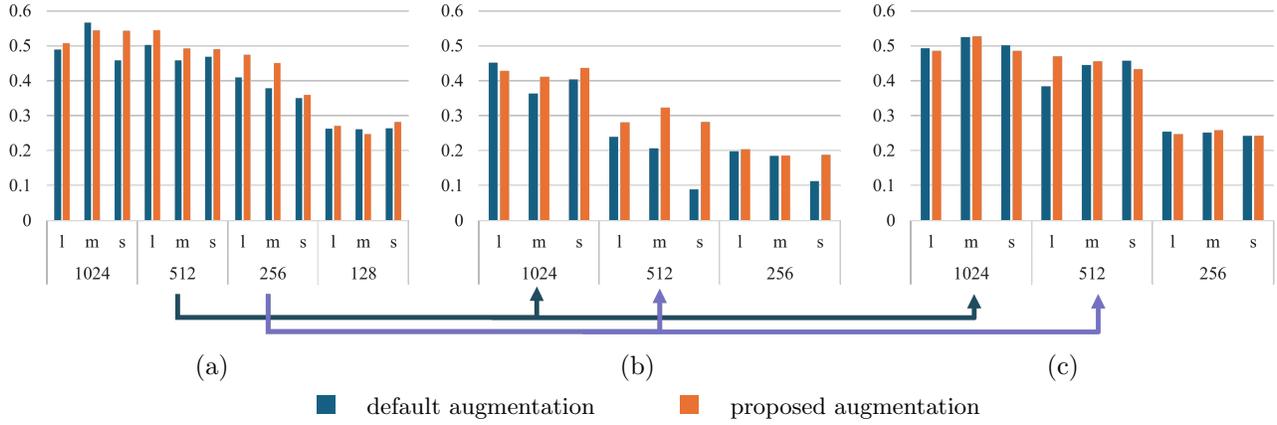
\begin{figure}[!ht]
    \centering
    \begin{minipage}{0.33\textwidth}
        \includegraphics[width=\linewidth]{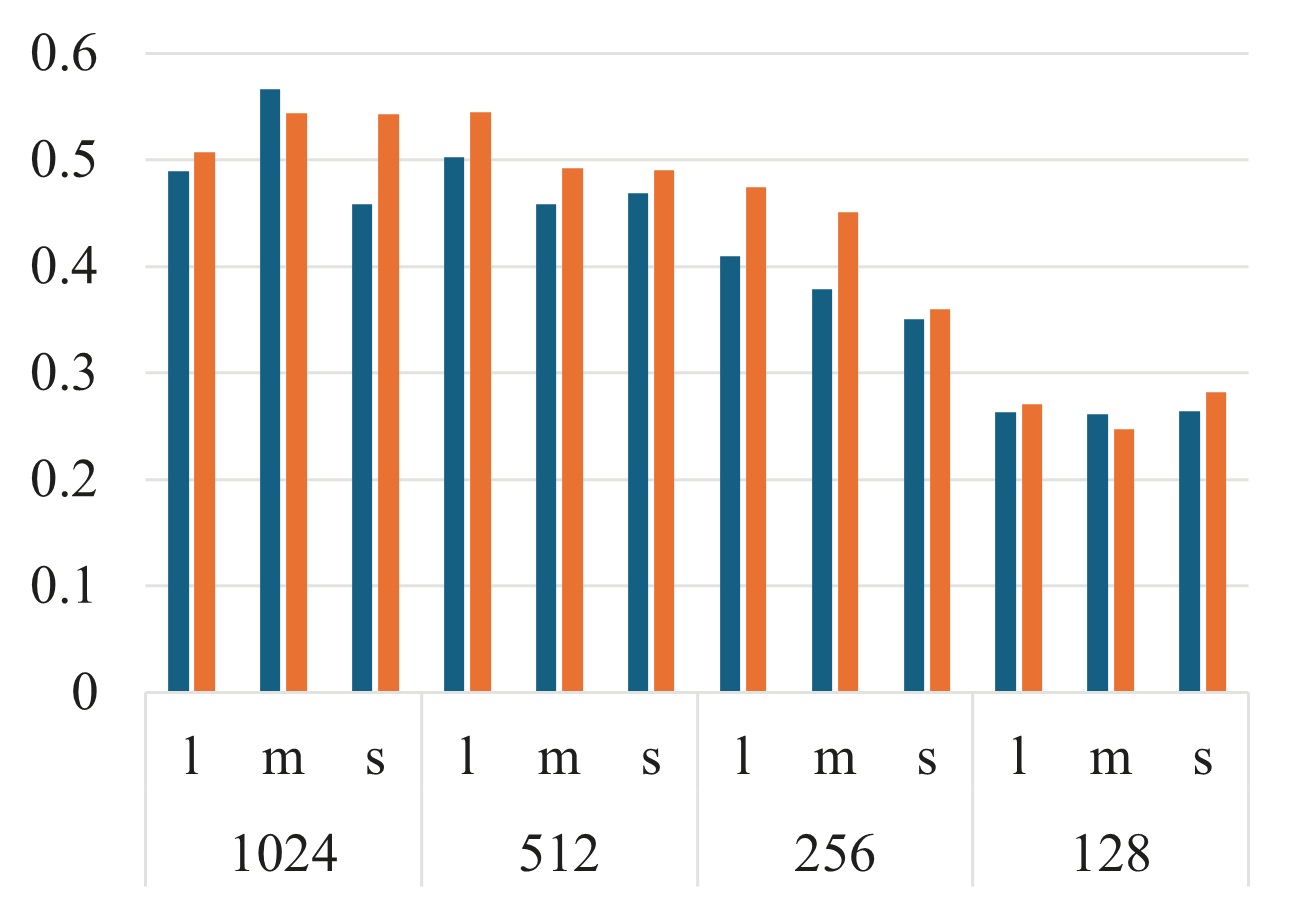}
    \end{minipage}%
    \hfill
    \begin{minipage}{0.33\textwidth}
        \includegraphics[width=\linewidth]{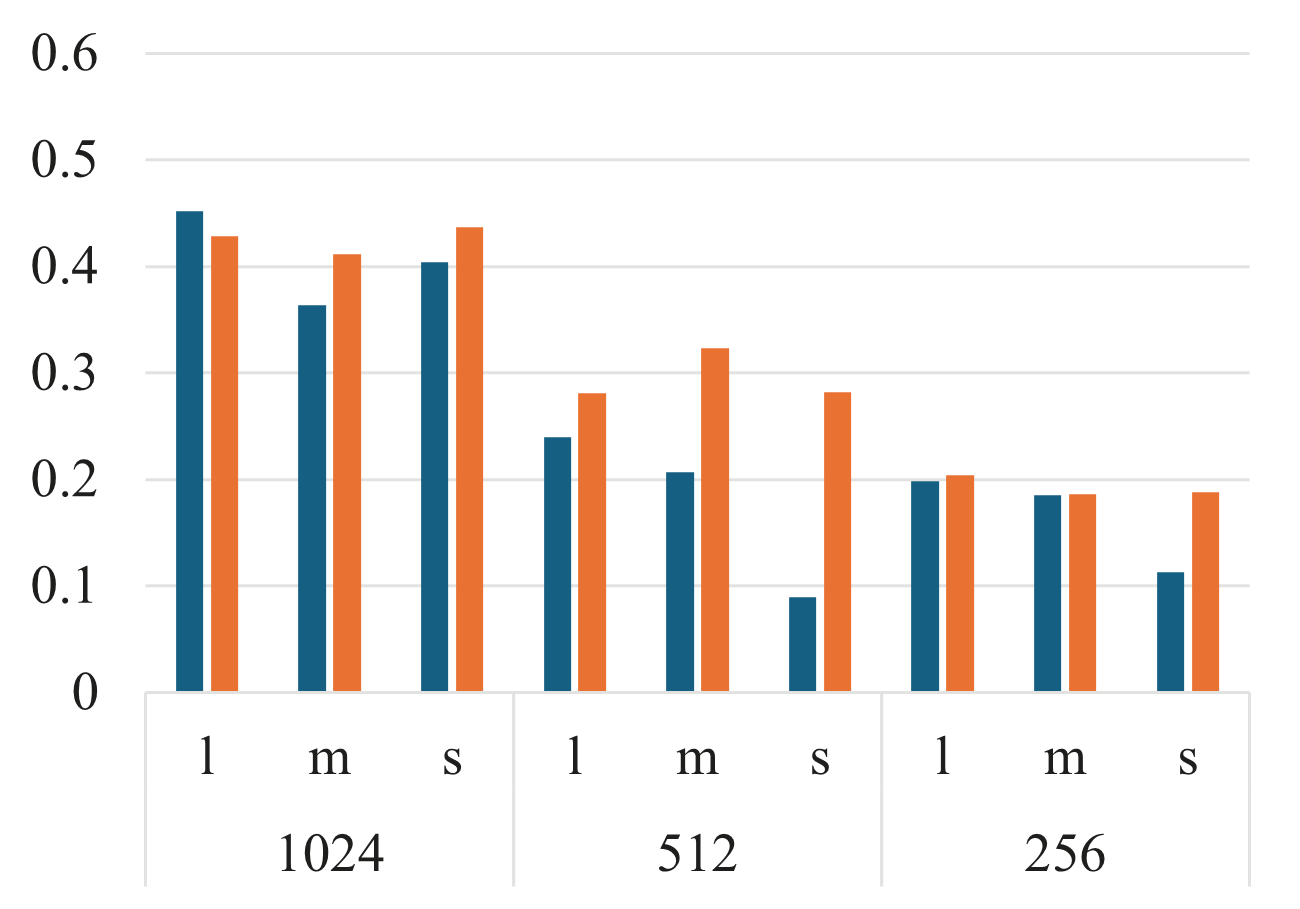}
    \end{minipage}%
    \hfill
    \begin{minipage}{0.33\textwidth}
        \includegraphics[width=\linewidth]{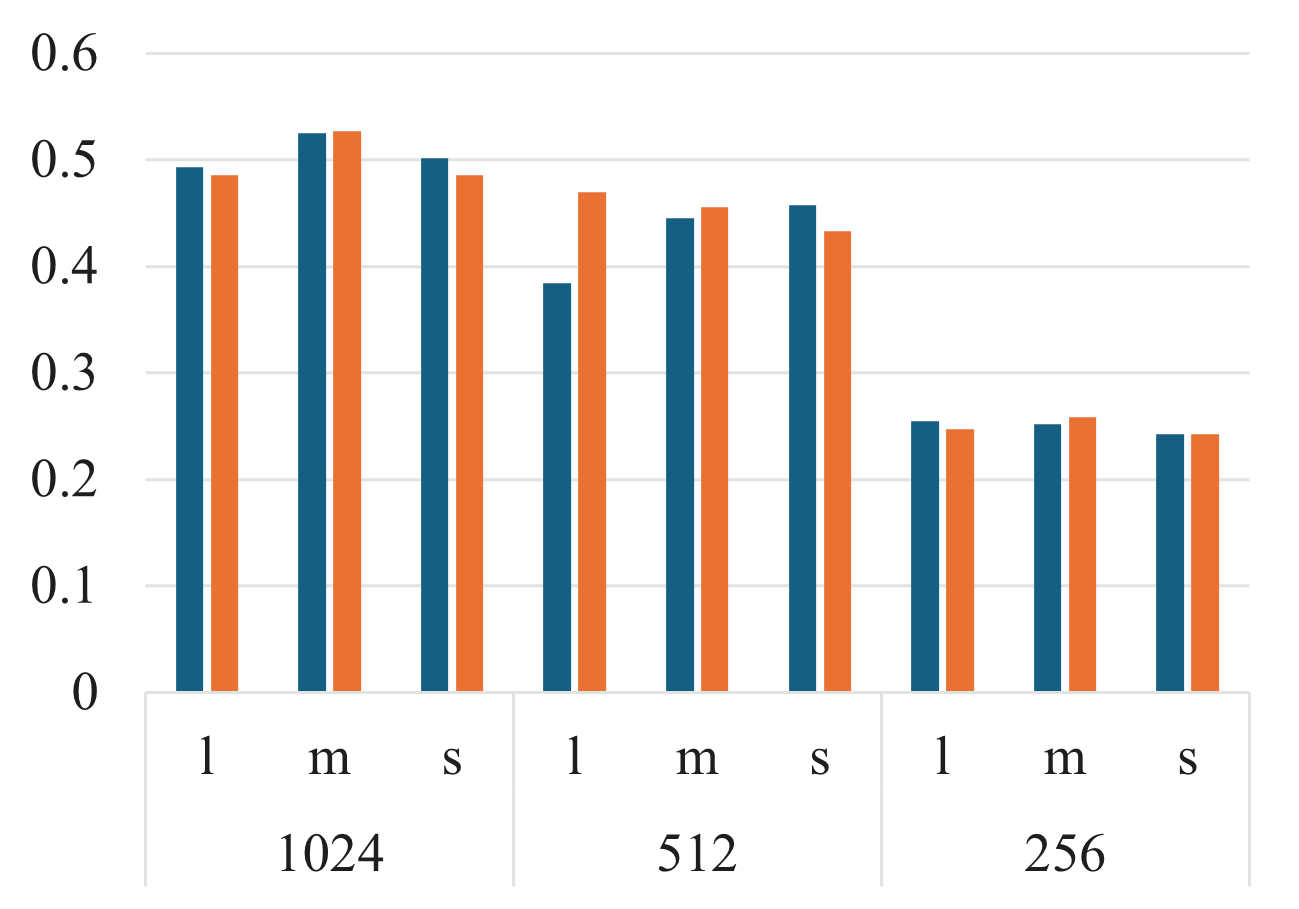}
    \end{minipage}%

    \vspace{-\baselineskip}
    \vspace{4pt}
    \begin{minipage}{\textwidth}
    \centering
        \includegraphics[width=0.726\linewidth]{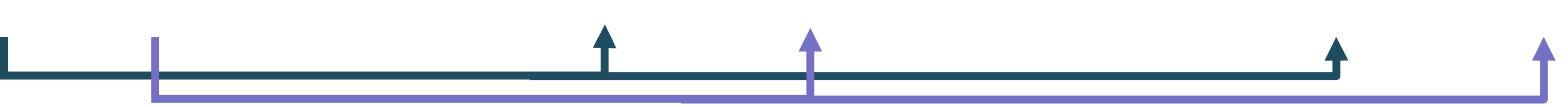}
    \end{minipage}%
    \vspace{6pt}
    \begin{minipage}{\linewidth}
        \begin{minipage}{0.33\linewidth}
        \centering
        (a)
        \end{minipage}%
        \begin{minipage}{0.33\linewidth}
        \centering
        (b)
        \end{minipage}%
        \begin{minipage}{0.33\linewidth}
        \centering
        (c)
        \end{minipage}%
    \end{minipage}%

    \begin{minipage}{\textwidth}        
            \smallskip
            \centering
            \begin{tabular}{c l @{\hspace{32pt}} c l}
                \vspace{16pt}
                \begin{tikzpicture}
                    \fill[custom_dark_blue] (0,0) rectangle (0.25,0.25);
                \end{tikzpicture} & \footnotesize \begin{tabular}[t]{@{}l@{}} default augmentation \\ \end{tabular} &
                
                \begin{tikzpicture}
                    \small
                    \fill[custom_orange] (0,0) rectangle (0.25,0.25);
                \end{tikzpicture} & \footnotesize \begin{tabular}[t]{@{}l@{}}  proposed augmentation \\ \end{tabular} \\
            \end{tabular}
        \end{minipage}
    \caption{(a) Baseline YOLO-NAS inference results (mAP @IoU0.5) on ADI-SEM dataset for different image  resolutions. [Ground-Truth image resolution: 1024, Downsampled (Bilinear interpolation) image resolutions: 128, 256, 512]. (b) Baseline YOLO-NAS inference results (mAP @IoU0.5) on ADI-SEM dataset for different image  resolutions, upsampled by Super-YOLOv5\cite{zhang2023superyolo} [512$\longrightarrow$ 1024, 256$\longrightarrow$ 512, 128$\longrightarrow$ 256]. (c) Baseline YOLO-NAS inference results (mAP @IoU0.5) on ADI-SEM dataset for different image  resolutions, upsampled by proposed SEMI-SuperYOLO-NAS  [512$\longrightarrow$ 1024, 256$\longrightarrow$ 512, 128$\longrightarrow$ 256]. **Shown for baseline YOLO-NAS model sizes as Small (S), Medium (M) and Large (L) with and without proposed augmentation strategy.}
    \label{SY5-SR}
\end{figure}

\begin{figure}[!ht]
\centering
\includegraphics[width=\linewidth]{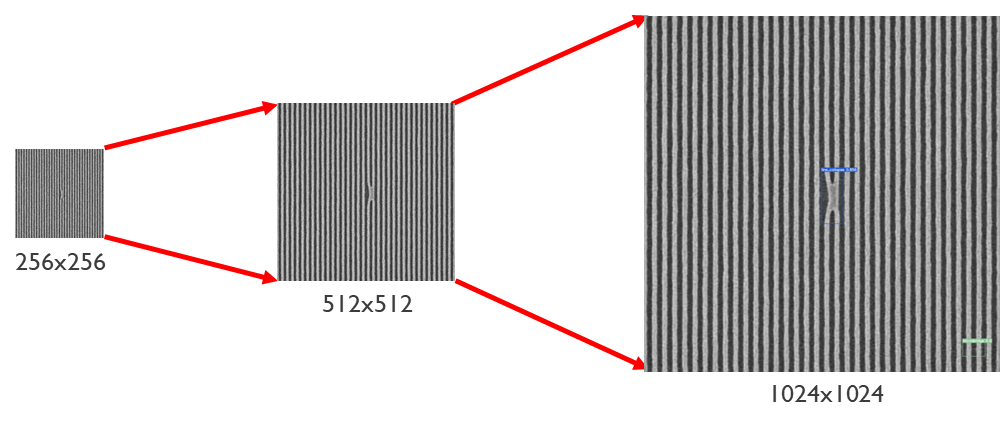}
\caption{Detection of defect instances using baseline YOLO-NAS on upsampled image resolution [256 --$\longrightarrow$ 1024] with the proposed SEMI-SuperYOLO-NAS SR branch.}
\label{SR_image}
\end{figure}

Subsequently, we incrementally upsampled each low-resolution (LR) image set (128, 256, and 512) by a factor of 2 ($\times$2) and generated/reconstructed corresponding high-resolution (HR) image sets (256, 512, and 1024). Initially, we applied the SR-assisted branch of the previous SuperYOLOv5\cite{zhang2023superyolo} and then the proposed SEMI-SuperYOLO-NAS. Our observations revealed that the detection performance on upsampled images (1024, 512, 256) using SuperYOLOv5\cite{zhang2023superyolo} SR-branch was comparatively lower than that on corresponding low-resolution images (512, 256, 128), as illustrated in Fig.\ref{SY5-SR} (b) against \ref{SY5-SR} (a). However, the detection performance on upsampled images (1024, 512, 256) using our proposed SEMI-SuperYOLO-NAS SR-branch was relatively higher or on par with corresponding low-resolution images (512, 256, 128), as depicted in Fig.\ref{SY5-SR}(c) against \ref{SY5-SR} (a). The performance of defect detection is illustrated in Fig.\ref{SR_image}.

\section{Limitations and Future Work}
The present study highlights several limitations and suggests avenues for future research based on the findings. Firstly, while the SR-branch employed improves intermediate high-resolution feature learning, it lacks optimization for high-quality SR image reconstruction or generation. One potential future direction involves training a novel SR-branch specifically tailored specifically for generating high-quality Super-Resolution SEM images. Moreover, the study reveals that the baseline YOLO-NAS struggles with detecting critically small defects, particularly gap and probable-gap anomalies. Further investigation is necessary to determine if the NAS space is restricted by limited computational resources, potentially leading to suboptimal results for specific defect classes. This is particularly pertinent considering that NAS was conducted using the Object365 dataset and COCO Pseudo labeled dataset, which possess different characteristics compared to semiconductor image datasets. Additionally, the utilization of low-resolution/downscale images obtained through Bilinear interpolation poses challenges, as it fails to effectively handle high-frequency details or sharp transitions in the image, resulting in image blurring, loss of fine details, and aliasing artifacts. Future endeavors could focus on employing more sophisticated algorithms such as Bicubic interpolation, Lanczos resampling, or advanced deep learning methods to address these challenges. Additionally, the acquisition of original wafer SEM images at various resolution settings (such as 128, 256, 512, 1024,...) could offer valuable data for further research and development in this field.

\section{Conclusion}

In this research, we introduce a novel architecture of Super-YOLO-NAS for semiconductor defect inspection, aimed at enhancing SEM throughput by reducing required SEM image pixel resolutions by approximately a factor of 8 ($\sim\times$8). Compared to the baseline YOLO-NAS, our proposed methodology enhances detection precision by 23\% for SEM-ADI and 9\% for EDR-AEI dataset, at lower SEM resolutions. Furthermore, we illustrate zero-shot inference on a novel SEM-ADI test dataset, obtained under distinct process and imaging conditions, diverging from the distribution of the training dataset. The proposed architecture demonstrates robustness and generalizability, proficiently capturing semantic embeddings for specific defects such as line collapse, bridging, and micro-bridging, even when the train and test datasets pertain to different process conditions, varied CD/pitch, and imaging conditions. Finally, we demonstrated the performance of the baseline YOLO-NAS inference on upsampled images, which were generated from their respective low-resolution counterparts using the SR-assisted branch of the proposed Super-YOLO-NAS. This evaluation was conducted by comparing against upsampled images generated by the SR-assisted branch of Super-YOLOv5, a previous framework based on the YOLOv5 model. 

\section*{Author Contribution}
\begin{itemize}
    \item Bappaditya Dey conceived the idea, planned, and led the project.
    \item Bappaditya Dey proposed the methodology in discussion with Ying-Lin Chen, Vic De Ridder, Jacob Deforce, Victor Blanco, and Sandip Halder.
    \item Ying-Lin Chen, Vic De Ridder and Jacob Deforce performed experiments with the advice of Bappaditya Dey.
    \item Ying-Lin Chen, Vic De Ridder and Jacob Deforce performed data analysis with the advice of Bappaditya Dey.
    \item Ying-Lin Chen, Vic De Ridder and Jacob Deforce wrote the manuscript.
    \item Bappaditya Dey reviewed and edited the final manuscript.
    \item All the authors contributed to the discussions on the results and reviewed the manuscript.
\end{itemize}

\bibliography{main} 
\bibliographystyle{spiebib} 

\end{document}